\documentclass{article}

\IfFileExists{neurips_2026.sty}%
  {\usepackage[final]{neurips_2026}}%
  {\usepackage[margin=1in]{geometry}\usepackage{natbib}%
   \newenvironment{ack}{\section*{Acknowledgments}}{}}
\makeatletter
\@ifpackageloaded{neurips_2026}{%
  \renewcommand{\@noticestring}{Workshop on Who Verifies the Agents? Toward Reliable Agent Development, NeurIPS 2026.}}{}
\makeatother

\usepackage[utf8]{inputenc}
\usepackage[T1]{fontenc}
\usepackage{microtype}
\usepackage{graphicx}
\usepackage{booktabs}
\usepackage{amsmath}
\usepackage{amssymb}
\usepackage{xcolor}
\usepackage{tikz}
\usetikzlibrary{positioning,arrows.meta,calc}
\usepackage{fontawesome5}
\usepackage{placeins}
\usepackage[hidelinks]{hyperref}
\hypersetup{pdftitle={trajectory-judge: What Outcome-Only LLM Judges Miss on Agent Trajectories},
  pdfauthor={Hadi Mohammadi}}

\newcommand{\tjRulesFone}{0.800}

\newcommand{\tjRulesSilent}{0.429}

\newcommand{\tjRulesLoud}{1.000}
\newcommand{\tjRulesLoudCI}{[0.97, 1.00]}
\newcommand{\tjRulesFA}{0.000}
\newcommand{\tjRulesFACI}{[0.00, 0.04]}
\newcommand{\tjRulesStepExact}{1.000}

\newcommand{\tjRulesTypeF}{0.667}

\newcommand{\tjRulesECE}{0.075}

\newcommand{\tjRulesWrongTool}{0.00}
\newcommand{\tjRulesWrongToolCI}{[0.00, 0.07]}
\newcommand{\tjRulesHallucArg}{1.00}
\newcommand{\tjRulesHallucArgCI}{[0.93, 1.00]}
\newcommand{\tjRulesSkippedPre}{1.00}
\newcommand{\tjRulesSkippedPreCI}{[0.93, 1.00]}
\newcommand{\tjRulesIgnoredObs}{1.00}
\newcommand{\tjRulesIgnoredObsCI}{[0.93, 1.00]}
\newcommand{\tjRulesPrematureStop}{1.00}
\newcommand{\tjRulesPrematureStopCI}{[0.93, 1.00]}
\newcommand{\tjRulesUnsupClaim}{0.00}
\newcommand{\tjRulesUnsupClaimCI}{[0.00, 0.07]}
\newcommand{\tjRulesFATwo}{0.00}
\newcommand{\tjRulesCost}{0.0}
\newcommand{\tjRulesStepN}{200}
\newcommand{\tjOutcomeFone}{0.712}
\newcommand{\tjOutcomeFoneCI}{[0.67, 0.75]}
\newcommand{\tjOutcomeSilent}{0.451}
\newcommand{\tjOutcomeSilentCI}{[0.38, 0.52]}
\newcommand{\tjOutcomeLoud}{0.840}
\newcommand{\tjOutcomeLoudCI}{[0.78, 0.90]}
\newcommand{\tjOutcomeFA}{0.330}
\newcommand{\tjOutcomeFACI}{[0.24, 0.43]}
\newcommand{\tjOutcomeTypeF}{0.300}
\newcommand{\tjOutcomeTypeFCI}{[0.27, 0.33]}
\newcommand{\tjOutcomeECE}{0.253}

\newcommand{\tjOutcomeWrongTool}{0.34}
\newcommand{\tjOutcomeWrongToolCI}{[0.22, 0.48]}
\newcommand{\tjOutcomeHallucArg}{0.34}
\newcommand{\tjOutcomeHallucArgCI}{[0.22, 0.48]}
\newcommand{\tjOutcomeSkippedPre}{0.74}
\newcommand{\tjOutcomeSkippedPreCI}{[0.62, 0.86]}
\newcommand{\tjOutcomeIgnoredObs}{0.76}
\newcommand{\tjOutcomeIgnoredObsCI}{[0.64, 0.88]}
\newcommand{\tjOutcomePrematureStop}{1.00}
\newcommand{\tjOutcomePrematureStopCI}{[0.93, 1.00]}
\newcommand{\tjOutcomeUnsupClaim}{0.50}
\newcommand{\tjOutcomeUnsupClaimCI}{[0.36, 0.64]}
\newcommand{\tjOutcomeFATwo}{0.33}
\newcommand{\tjOutcomeCost}{3.5}

\newcommand{\tjStepQFone}{0.923}
\newcommand{\tjStepQFoneCI}{[0.91, 0.93]}
\newcommand{\tjStepQSilent}{0.766}
\newcommand{\tjStepQSilentCI}{[0.74, 0.80]}
\newcommand{\tjStepQLoud}{0.984}
\newcommand{\tjStepQLoudCI}{[0.96, 1.00]}
\newcommand{\tjStepQFA}{0.000}
\newcommand{\tjStepQFACI}{[0.00, 0.04]}
\newcommand{\tjStepQStepExact}{0.973}

\newcommand{\tjStepQTypeF}{0.606}
\newcommand{\tjStepQTypeFCI}{[0.55, 0.65]}
\newcommand{\tjStepQECE}{0.033}

\newcommand{\tjStepQWrongTool}{1.00}
\newcommand{\tjStepQWrongToolCI}{[0.93, 1.00]}
\newcommand{\tjStepQHallucArg}{1.00}
\newcommand{\tjStepQHallucArgCI}{[0.93, 1.00]}
\newcommand{\tjStepQSkippedPre}{1.00}
\newcommand{\tjStepQSkippedPreCI}{[0.93, 1.00]}
\newcommand{\tjStepQIgnoredObs}{1.00}
\newcommand{\tjStepQIgnoredObsCI}{[0.93, 1.00]}
\newcommand{\tjStepQPrematureStop}{0.96}
\newcommand{\tjStepQPrematureStopCI}{[0.90, 1.00]}
\newcommand{\tjStepQUnsupClaim}{0.18}
\newcommand{\tjStepQUnsupClaimCI}{[0.08, 0.30]}
\newcommand{\tjStepQFATwo}{0.00}
\newcommand{\tjStepQCost}{10.4}
\newcommand{\tjStepQStepN}{223}
\newcommand{\tjStepLFone}{0.852}
\newcommand{\tjStepLFoneCI}{[0.85, 0.86]}
\newcommand{\tjStepLSilent}{0.983}
\newcommand{\tjStepLSilentCI}{[0.96, 1.00]}
\newcommand{\tjStepLLoud}{1.000}
\newcommand{\tjStepLLoudCI}{[0.97, 1.00]}
\newcommand{\tjStepLFA}{1.000}
\newcommand{\tjStepLFACI}{[0.96, 1.00]}
\newcommand{\tjStepLStepExact}{0.807}

\newcommand{\tjStepLTypeF}{0.311}
\newcommand{\tjStepLTypeFCI}{[0.27, 0.35]}
\newcommand{\tjStepLECE}{0.148}

\newcommand{\tjStepLWrongTool}{1.00}
\newcommand{\tjStepLWrongToolCI}{[0.93, 1.00]}
\newcommand{\tjStepLHallucArg}{0.94}
\newcommand{\tjStepLHallucArgCI}{[0.86, 1.00]}
\newcommand{\tjStepLSkippedPre}{1.00}
\newcommand{\tjStepLSkippedPreCI}{[0.93, 1.00]}
\newcommand{\tjStepLIgnoredObs}{1.00}
\newcommand{\tjStepLIgnoredObsCI}{[0.93, 1.00]}
\newcommand{\tjStepLPrematureStop}{1.00}
\newcommand{\tjStepLPrematureStopCI}{[0.93, 1.00]}
\newcommand{\tjStepLUnsupClaim}{1.00}
\newcommand{\tjStepLUnsupClaimCI}{[0.93, 1.00]}
\newcommand{\tjStepLFATwo}{1.00}
\newcommand{\tjStepLCost}{1.5}
\newcommand{\tjStepLStepN}{275}
\newcommand{\tjSelfconsFone}{0.913}
\newcommand{\tjSelfconsFoneCI}{[0.90, 0.93]}
\newcommand{\tjSelfconsSilent}{0.760}
\newcommand{\tjSelfconsSilentCI}{[0.73, 0.79]}
\newcommand{\tjSelfconsLoud}{0.960}
\newcommand{\tjSelfconsLoudCI}{[0.92, 0.99]}
\newcommand{\tjSelfconsFA}{0.010}
\newcommand{\tjSelfconsFACI}{[0.00, 0.03]}
\newcommand{\tjSelfconsStepExact}{0.982}

\newcommand{\tjSelfconsTypeF}{0.583}
\newcommand{\tjSelfconsTypeFCI}{[0.53, 0.63]}
\newcommand{\tjSelfconsECE}{0.084}

\newcommand{\tjSelfconsWrongTool}{1.00}
\newcommand{\tjSelfconsWrongToolCI}{[0.93, 1.00]}
\newcommand{\tjSelfconsHallucArg}{1.00}
\newcommand{\tjSelfconsHallucArgCI}{[0.93, 1.00]}
\newcommand{\tjSelfconsSkippedPre}{1.00}
\newcommand{\tjSelfconsSkippedPreCI}{[0.93, 1.00]}
\newcommand{\tjSelfconsIgnoredObs}{1.00}
\newcommand{\tjSelfconsIgnoredObsCI}{[0.93, 1.00]}
\newcommand{\tjSelfconsPrematureStop}{0.90}
\newcommand{\tjSelfconsPrematureStopCI}{[0.80, 0.98]}
\newcommand{\tjSelfconsUnsupClaim}{0.16}
\newcommand{\tjSelfconsUnsupClaimCI}{[0.06, 0.26]}
\newcommand{\tjSelfconsFATwo}{0.01}
\newcommand{\tjSelfconsCost}{30.2}
\newcommand{\tjSelfconsStepN}{228}
\newcommand{\tjDeltaOutcomeLoudSilent}{+0.389}
\newcommand{\tjDeltaOutcomeLoudSilentCI}{[+0.296, +0.482]}
\newcommand{\tjDeltaStepOutcomeSilent}{+0.314}
\newcommand{\tjDeltaStepOutcomeSilentCI}{[+0.240, +0.389]}
\newcommand{\tjDeltaStepRulesSilent}{+0.337}
\newcommand{\tjDeltaStepRulesSilentCI}{[+0.309, +0.371]}
\newcommand{\tjDeltaSelfconsFone}{\ensuremath{-}0.009}
\newcommand{\tjDeltaSelfconsFoneCI}{[\ensuremath{-}0.020, +0.000]}
\newcommand{\tjDeltaSelfconsSilent}{\ensuremath{-}0.006}
\newcommand{\tjDeltaSelfconsSilentCI}{[\ensuremath{-}0.017, +0.000]}
\newcommand{\tjDeltaSelfconsTypeF}{\ensuremath{-}0.023}
\newcommand{\tjDeltaSelfconsTypeFCI}{[\ensuremath{-}0.079, +0.035]}
\newcommand{\tjDeltaSelfconsECE}{+0.051}
\newcommand{\tjDeltaSelfconsECECI}{[+0.039, +0.071]}
\newcommand{\tjDeltaSelfconsStepExact}{+0.009}
\newcommand{\tjDeltaSelfconsStepExactCI}{[\ensuremath{-}0.008, +0.027]}

\newcommand{\tjStepQFlagsTrue}{257}
\newcommand{\tjStepLFlags}{397}

\newcommand{\tjSelfconsTwoThirds}{43}
\newcommand{\tjSelfconsUnanimous}{357}
\newcommand{\tjStepQMeanTokens}{391}
\newcommand{\tjStepLMeanTokens}{57}
\newcommand{\tjNTraj}{400}
\newcommand{\tjNClean}{100}
\newcommand{\tjNFaulty}{300}
\newcommand{\tjNSilent}{175}
\newcommand{\tjNLoud}{125}
\newcommand{\tjNPerType}{50}
\newcommand{\tjNVerdicts}{2000}

\newcommand{\tjNRestocking}{128}
\newcommand{\tjAgentN}{60}
\newcommand{\tjAgentFaulty}{13}
\newcommand{\tjAgentWrongOutcome}{10}
\newcommand{\tjAgentSilent}{3}

\newcommand{\tjOutcomeLoudPct}{84\%}
\newcommand{\tjOutcomeSilentPct}{45\%}

\newcommand{\tjCaseConf}{0.92}

\newcommand{\tjStepQHallucAsWrong}{35}
\newcommand{\tjStepQUnsupMissed}{41}

\newcommand{\tjStepQPremCorrect}{14}
\newcommand{\tjStepQPremAsUnsup}{17}
\newcommand{\tjStepQPremAsSkipped}{11}
\newcommand{\tjRulesStepWithinOne}{1.000}
\newcommand{\tjStepQStepWithinOne}{0.982}
\newcommand{\tjStepLStepWithinOne}{0.862}
\newcommand{\tjSelfconsStepWithinOne}{0.987}
\newcommand{\tjRulesPromptTok}{0}
\newcommand{\tjRulesComplTok}{0}
\newcommand{\tjOutcomePromptTok}{231,320}
\newcommand{\tjOutcomeComplTok}{52,855}
\newcommand{\tjStepQPromptTok}{405,212}
\newcommand{\tjStepQComplTok}{156,412}
\newcommand{\tjStepLPromptTok}{380,706}
\newcommand{\tjStepLComplTok}{22,754}
\newcommand{\tjSelfconsPromptTok}{1,215,636}
\newcommand{\tjSelfconsComplTok}{486,207}
\newcommand{\tjTotalHours}{5.05}
\newcommand{\tjAgentAlreadyRefundedFaulty}{10}
\newcommand{\tjAgentWrongCustomerFaulty}{3}
\newcommand{\tjNReplySame}{200}
\newcommand{\tjNReplySameKept}{125}
\newcommand{\tjNReplySameBroke}{75}
\newcommand{\tjNReplyChanged}{100}
\newcommand{\tjNPairs}{251}
\newcommand{\tjNPairsReplySame}{151}
\newcommand{\tjNPairsReplyChanged}{100}
\newcommand{\tjNPairsLoud}{84}
\newcommand{\tjNPairsSilent}{167}
\newcommand{\tjNLatePairs}{49}
\newcommand{\tjNLateParents}{41}
\newcommand{\tjOutcomeIdenticalPairs}{151}
\newcommand{\tjOutcomeFAHappyK}{13}
\newcommand{\tjOutcomeFAHappyN}{17}
\newcommand{\tjOutcomeFARestockK}{10}
\newcommand{\tjOutcomeFARestockN}{17}
\newcommand{\tjOutcomeFAAlreadyK}{10}
\newcommand{\tjOutcomeFAAlreadyN}{16}
\newcommand{\tjOutcomeFAEscalK}{0}
\newcommand{\tjOutcomeFAEscalN}{50}
\newcommand{\tjOutcomeSameHappyK}{34}
\newcommand{\tjOutcomeSameHappyN}{43}
\newcommand{\tjOutcomeSameRestockK}{67}
\newcommand{\tjOutcomeSameRestockN}{93}
\newcommand{\tjOutcomeSameAlreadyK}{8}
\newcommand{\tjOutcomeSameAlreadyN}{16}
\newcommand{\tjOutcomeSameEscalK}{0}
\newcommand{\tjOutcomeSameEscalN}{48}
\newcommand{\tjOutcomeFAStratumMaxPct}{76\%}
\newcommand{\tjOutcomeFAStratumMinPct}{0\%}
\newcommand{\tjOutcomeReplySameRecallMinPct}{34\%}
\newcommand{\tjOutcomeReplySameRecallMaxPct}{76\%}
\newcommand{\tjOutcomeLoudFlags}{105}
\newcommand{\tjOutcomeLoudFlagsSame}{55}
\newcommand{\tjRulesPairWrongTool}{+0.00}
\newcommand{\tjRulesPairWrongToolDisc}{0/0}
\newcommand{\tjRulesPairWrongToolP}{---}

\newcommand{\tjRulesPairHallucArg}{+1.00}
\newcommand{\tjRulesPairHallucArgDisc}{50/0}
\newcommand{\tjRulesPairHallucArgP}{<0.001}

\newcommand{\tjRulesPairSkippedPre}{+1.00}
\newcommand{\tjRulesPairSkippedPreDisc}{34/0}
\newcommand{\tjRulesPairSkippedPreP}{<0.001}

\newcommand{\tjRulesPairIgnoredObs}{+1.00}
\newcommand{\tjRulesPairIgnoredObsDisc}{17/0}
\newcommand{\tjRulesPairIgnoredObsP}{<0.001}

\newcommand{\tjRulesPairPrematureStop}{+1.00}
\newcommand{\tjRulesPairPrematureStopDisc}{50/0}
\newcommand{\tjRulesPairPrematureStopP}{<0.001}

\newcommand{\tjRulesPairUnsupClaim}{+0.00}
\newcommand{\tjRulesPairUnsupClaimDisc}{0/0}
\newcommand{\tjRulesPairUnsupClaimP}{---}

\newcommand{\tjRulesPairReplySame}{+0.669}
\newcommand{\tjRulesPairReplySameDisc}{101/0}
\newcommand{\tjRulesPairReplySameCI}{[+0.63, +0.71]}
\newcommand{\tjRulesPairReplyChanged}{+0.500}
\newcommand{\tjRulesPairReplyChangedDisc}{50/0}
\newcommand{\tjRulesPairReplyChangedCI}{[+0.50, +0.50]}

\newcommand{\tjRulesPairAll}{+0.602}

\newcommand{\tjRulesPairAllCI}{[+0.58, +0.63]}
\newcommand{\tjRulesPairReplySameKept}{+0.57}

\newcommand{\tjRulesReplySameKept}{0.60}
\newcommand{\tjRulesPairReplySameBroke}{+1.00}

\newcommand{\tjRulesReplySameBroke}{1.00}

\newcommand{\tjOutcomePairWrongTool}{+0.00}
\newcommand{\tjOutcomePairWrongToolDisc}{0/0}
\newcommand{\tjOutcomePairWrongToolP}{---}

\newcommand{\tjOutcomePairHallucArg}{+0.00}
\newcommand{\tjOutcomePairHallucArgDisc}{0/0}
\newcommand{\tjOutcomePairHallucArgP}{---}

\newcommand{\tjOutcomePairSkippedPre}{+0.00}
\newcommand{\tjOutcomePairSkippedPreDisc}{0/0}
\newcommand{\tjOutcomePairSkippedPreP}{---}

\newcommand{\tjOutcomePairIgnoredObs}{+0.00}
\newcommand{\tjOutcomePairIgnoredObsDisc}{0/0}
\newcommand{\tjOutcomePairIgnoredObsP}{---}

\newcommand{\tjOutcomePairPrematureStop}{+0.66}
\newcommand{\tjOutcomePairPrematureStopDisc}{33/0}
\newcommand{\tjOutcomePairPrematureStopP}{<0.001}

\newcommand{\tjOutcomePairUnsupClaim}{+0.16}
\newcommand{\tjOutcomePairUnsupClaimDisc}{10/2}
\newcommand{\tjOutcomePairUnsupClaimP}{0.039}

\newcommand{\tjOutcomePairReplySame}{+0.000}
\newcommand{\tjOutcomePairReplySameDisc}{0/0}

\newcommand{\tjOutcomePairReplyChanged}{+0.410}
\newcommand{\tjOutcomePairReplyChangedDisc}{43/2}
\newcommand{\tjOutcomePairReplyChangedCI}{[+0.30, +0.52]}
\newcommand{\tjOutcomePairLoud}{+0.393}

\newcommand{\tjOutcomePairSilent}{+0.048}

\newcommand{\tjOutcomePairAll}{+0.163}

\newcommand{\tjOutcomePairAllCI}{[+0.12, +0.21]}
\newcommand{\tjOutcomePairReplySameKept}{+0.00}

\newcommand{\tjOutcomeReplySameKept}{0.43}
\newcommand{\tjOutcomePairReplySameBroke}{+0.00}

\newcommand{\tjOutcomeReplySameBroke}{0.73}
\newcommand{\tjOutcomeFAK}{33}
\newcommand{\tjStepQPairWrongTool}{+1.00}
\newcommand{\tjStepQPairWrongToolDisc}{50/0}
\newcommand{\tjStepQPairWrongToolP}{<0.001}

\newcommand{\tjStepQPairHallucArg}{+1.00}
\newcommand{\tjStepQPairHallucArgDisc}{50/0}
\newcommand{\tjStepQPairHallucArgP}{<0.001}

\newcommand{\tjStepQPairSkippedPre}{+1.00}
\newcommand{\tjStepQPairSkippedPreDisc}{34/0}
\newcommand{\tjStepQPairSkippedPreP}{<0.001}

\newcommand{\tjStepQPairIgnoredObs}{+1.00}
\newcommand{\tjStepQPairIgnoredObsDisc}{17/0}
\newcommand{\tjStepQPairIgnoredObsP}{<0.001}

\newcommand{\tjStepQPairPrematureStop}{+0.96}
\newcommand{\tjStepQPairPrematureStopDisc}{48/0}
\newcommand{\tjStepQPairPrematureStopP}{<0.001}

\newcommand{\tjStepQPairUnsupClaim}{+0.18}
\newcommand{\tjStepQPairUnsupClaimDisc}{9/0}
\newcommand{\tjStepQPairUnsupClaimP}{0.004}

\newcommand{\tjStepQPairReplySame}{+1.000}
\newcommand{\tjStepQPairReplySameDisc}{151/0}
\newcommand{\tjStepQPairReplySameCI}{[+1.00, +1.00]}
\newcommand{\tjStepQPairReplyChanged}{+0.570}
\newcommand{\tjStepQPairReplyChangedDisc}{57/0}
\newcommand{\tjStepQPairReplyChangedCI}{[+0.51, +0.63]}

\newcommand{\tjStepQPairAll}{+0.829}

\newcommand{\tjStepQPairAllCI}{[+0.80, +0.86]}
\newcommand{\tjStepQPairReplySameKept}{+1.00}

\newcommand{\tjStepQReplySameKept}{1.00}
\newcommand{\tjStepQPairReplySameBroke}{+1.00}

\newcommand{\tjStepQReplySameBroke}{1.00}
\newcommand{\tjStepQFAK}{0}
\newcommand{\tjStepLPairWrongTool}{+0.00}
\newcommand{\tjStepLPairWrongToolDisc}{0/0}
\newcommand{\tjStepLPairWrongToolP}{---}

\newcommand{\tjStepLPairHallucArg}{\ensuremath{-}0.06}
\newcommand{\tjStepLPairHallucArgDisc}{0/3}
\newcommand{\tjStepLPairHallucArgP}{0.250}

\newcommand{\tjStepLPairSkippedPre}{+0.00}
\newcommand{\tjStepLPairSkippedPreDisc}{0/0}
\newcommand{\tjStepLPairSkippedPreP}{---}

\newcommand{\tjStepLPairIgnoredObs}{+0.00}
\newcommand{\tjStepLPairIgnoredObsDisc}{0/0}
\newcommand{\tjStepLPairIgnoredObsP}{---}

\newcommand{\tjStepLPairPrematureStop}{+0.00}
\newcommand{\tjStepLPairPrematureStopDisc}{0/0}
\newcommand{\tjStepLPairPrematureStopP}{---}

\newcommand{\tjStepLPairUnsupClaim}{+0.00}
\newcommand{\tjStepLPairUnsupClaimDisc}{0/0}
\newcommand{\tjStepLPairUnsupClaimP}{---}

\newcommand{\tjStepLPairReplySame}{\ensuremath{-}0.020}
\newcommand{\tjStepLPairReplySameDisc}{0/3}
\newcommand{\tjStepLPairReplySameCI}{[\ensuremath{-}0.05, +0.00]}
\newcommand{\tjStepLPairReplyChanged}{+0.000}
\newcommand{\tjStepLPairReplyChangedDisc}{0/0}
\newcommand{\tjStepLPairReplyChangedCI}{---}

\newcommand{\tjStepLPairAll}{\ensuremath{-}0.012}

\newcommand{\tjStepLPairAllCI}{[\ensuremath{-}0.03, +0.00]}
\newcommand{\tjStepLPairReplySameKept}{\ensuremath{-}0.03}

\newcommand{\tjStepLReplySameKept}{0.98}
\newcommand{\tjStepLPairReplySameBroke}{+0.00}

\newcommand{\tjStepLReplySameBroke}{1.00}

\newcommand{\tjSelfconsPairWrongTool}{+1.00}
\newcommand{\tjSelfconsPairWrongToolDisc}{50/0}
\newcommand{\tjSelfconsPairWrongToolP}{<0.001}

\newcommand{\tjSelfconsPairHallucArg}{+1.00}
\newcommand{\tjSelfconsPairHallucArgDisc}{50/0}
\newcommand{\tjSelfconsPairHallucArgP}{<0.001}

\newcommand{\tjSelfconsPairSkippedPre}{+1.00}
\newcommand{\tjSelfconsPairSkippedPreDisc}{34/0}
\newcommand{\tjSelfconsPairSkippedPreP}{<0.001}

\newcommand{\tjSelfconsPairIgnoredObs}{+1.00}
\newcommand{\tjSelfconsPairIgnoredObsDisc}{17/0}
\newcommand{\tjSelfconsPairIgnoredObsP}{<0.001}

\newcommand{\tjSelfconsPairPrematureStop}{+0.90}
\newcommand{\tjSelfconsPairPrematureStopDisc}{45/0}
\newcommand{\tjSelfconsPairPrematureStopP}{<0.001}

\newcommand{\tjSelfconsPairUnsupClaim}{+0.16}
\newcommand{\tjSelfconsPairUnsupClaimDisc}{8/0}
\newcommand{\tjSelfconsPairUnsupClaimP}{0.008}

\newcommand{\tjSelfconsPairReplySame}{+1.000}
\newcommand{\tjSelfconsPairReplySameDisc}{151/0}
\newcommand{\tjSelfconsPairReplySameCI}{[+1.00, +1.00]}
\newcommand{\tjSelfconsPairReplyChanged}{+0.530}
\newcommand{\tjSelfconsPairReplyChangedDisc}{53/0}
\newcommand{\tjSelfconsPairReplyChangedCI}{[+0.46, +0.60]}

\newcommand{\tjSelfconsPairAll}{+0.813}

\newcommand{\tjSelfconsPairAllCI}{[+0.78, +0.84]}
\newcommand{\tjSelfconsPairReplySameKept}{+1.00}

\newcommand{\tjSelfconsReplySameKept}{1.00}
\newcommand{\tjSelfconsPairReplySameBroke}{+1.00}

\newcommand{\tjSelfconsReplySameBroke}{1.00}

\newcommand{\tjPairNWrongTool}{50}
\newcommand{\tjPairNHallucArg}{50}
\newcommand{\tjPairNSkippedPre}{34}
\newcommand{\tjPairNIgnoredObs}{17}
\newcommand{\tjPairNPrematureStop}{50}
\newcommand{\tjPairNUnsupClaim}{50}
\newcommand{\tjNPairsReplySameKept}{117}
\newcommand{\tjNPairsReplySameBroke}{34}
\newcommand{\tjOutcomePairGap}{+0.345}
\newcommand{\tjOutcomePairGapCI}{[+0.240, +0.467]}
\newcommand{\tjDeltaPairUnsupStepOutcome}{+0.020}
\newcommand{\tjDeltaPairUnsupStepOutcomeCI}{[\ensuremath{-}0.111, +0.152]}
\newcommand{\tjRulesSens}{0.667}

\newcommand{\tjRulesSpec}{1.000}

\newcommand{\tjRulesTypeJoint}{0.667}

\newcommand{\tjRulesBrierFaulty}{0.122}

\newcommand{\tjRulesBrierClean}{0.160}

\newcommand{\tjRulesPPVFive}{1.00}

\newcommand{\tjRulesLocDet}{1.000}

\newcommand{\tjRulesLocJoint}{0.667}

\newcommand{\tjRulesLocJointNonOmit}{0.600}

\newcommand{\tjRulesLocNonOmitK}{150}
\newcommand{\tjRulesLocNonOmitN}{150}

\newcommand{\tjOutcomeSens}{0.613}
\newcommand{\tjOutcomeSensCI}{[0.57, 0.66]}
\newcommand{\tjOutcomeSpec}{0.670}
\newcommand{\tjOutcomeSpecCI}{[0.57, 0.76]}

\newcommand{\tjOutcomeTypeJoint}{0.330}
\newcommand{\tjOutcomeTypeJointCI}{[0.30, 0.36]}
\newcommand{\tjOutcomeBrierFaulty}{0.290}
\newcommand{\tjOutcomeBrierFaultyCI}{[0.255, 0.325]}
\newcommand{\tjOutcomeBrierClean}{0.274}
\newcommand{\tjOutcomeBrierCleanCI}{[0.203, 0.346]}
\newcommand{\tjOutcomePPVFive}{0.09}
\newcommand{\tjOutcomePPVFiveCI}{[0.07, 0.12]}
\newcommand{\tjStepQSens}{0.857}
\newcommand{\tjStepQSensCI}{[0.84, 0.88]}
\newcommand{\tjStepQSpec}{1.000}
\newcommand{\tjStepQSpecCI}{---}

\newcommand{\tjStepQTypeJoint}{0.610}
\newcommand{\tjStepQTypeJointCI}{[0.57, 0.65]}
\newcommand{\tjStepQBrierFaulty}{0.128}
\newcommand{\tjStepQBrierFaultyCI}{[0.111, 0.144]}
\newcommand{\tjStepQBrierClean}{0.006}
\newcommand{\tjStepQBrierCleanCI}{[0.006, 0.006]}
\newcommand{\tjStepQPPVFive}{1.00}

\newcommand{\tjStepQLocDet}{0.844}
\newcommand{\tjStepQLocDetCI}{[0.82, 0.87]}
\newcommand{\tjStepQLocJoint}{0.723}
\newcommand{\tjStepQLocJointCI}{[0.70, 0.75]}
\newcommand{\tjStepQLocJointNonOmit}{0.836}

\newcommand{\tjStepQLocInvalid}{34}
\newcommand{\tjStepQLocNonOmitK}{209}
\newcommand{\tjStepQLocNonOmitN}{209}

\newcommand{\tjStepLSens}{0.990}
\newcommand{\tjStepLSensCI}{[0.98, 1.00]}
\newcommand{\tjStepLSpec}{0.000}
\newcommand{\tjStepLSpecCI}{---}

\newcommand{\tjStepLTypeJoint}{0.400}
\newcommand{\tjStepLTypeJointCI}{[0.37, 0.43]}
\newcommand{\tjStepLBrierFaulty}{0.021}
\newcommand{\tjStepLBrierFaultyCI}{[0.012, 0.032]}
\newcommand{\tjStepLBrierClean}{0.776}
\newcommand{\tjStepLBrierCleanCI}{[0.762, 0.790]}
\newcommand{\tjStepLPPVFive}{0.05}

\newcommand{\tjStepLLocDet}{0.747}
\newcommand{\tjStepLLocDetCI}{[0.71, 0.78]}
\newcommand{\tjStepLLocJoint}{0.740}
\newcommand{\tjStepLLocJointCI}{[0.71, 0.77]}
\newcommand{\tjStepLLocJointNonOmit}{0.776}

\newcommand{\tjStepLLocNonOmitK}{194}
\newcommand{\tjStepLLocNonOmitN}{247}

\newcommand{\tjSelfconsSens}{0.843}
\newcommand{\tjSelfconsSensCI}{[0.82, 0.87]}
\newcommand{\tjSelfconsSpec}{0.990}
\newcommand{\tjSelfconsSpecCI}{[0.97, 1.00]}

\newcommand{\tjSelfconsTypeJoint}{0.597}
\newcommand{\tjSelfconsTypeJointCI}{[0.56, 0.63]}
\newcommand{\tjSelfconsBrierFaulty}{0.139}
\newcommand{\tjSelfconsBrierFaultyCI}{[0.120, 0.156]}
\newcommand{\tjSelfconsBrierClean}{0.006}
\newcommand{\tjSelfconsBrierCleanCI}{[0.000, 0.016]}
\newcommand{\tjSelfconsPPVFive}{0.82}
\newcommand{\tjSelfconsPPVFiveCI}{[0.59, 1.00]}
\newcommand{\tjSelfconsLocDet}{0.885}
\newcommand{\tjSelfconsLocDetCI}{[0.86, 0.91]}
\newcommand{\tjSelfconsLocJoint}{0.747}
\newcommand{\tjSelfconsLocJointCI}{[0.72, 0.77]}
\newcommand{\tjSelfconsLocJointNonOmit}{0.832}

\newcommand{\tjSelfconsLocNonOmitK}{208}
\newcommand{\tjSelfconsLocNonOmitN}{208}

\newcommand{\tjRulesPPVFiveLow}{0.49}
\newcommand{\tjStepQPPVFiveLow}{0.55}
\newcommand{\tjStepQMistyped}{74}
\newcommand{\tjStepQMistypedPct}{29\%}
\newcommand{\tjDeltaSelfconsBrier}{+0.008}
\newcommand{\tjDeltaSelfconsBrierCI}{[\ensuremath{-}0.003, +0.019]}
\newcommand{\tjDeltaOutcomeStepBrier}{+0.189}
\newcommand{\tjDeltaOutcomeStepBrierCI}{[+0.156, +0.222]}
\newcommand{\tjDeltaOutcomeStepECE}{+0.220}
\newcommand{\tjDeltaOutcomeStepECECI}{[+0.178, +0.264]}
\newcommand{\tjSelfconsUnsupMissed}{42}
\newcommand{\tjSelfconsUnsupMissUnanimous}{32}
\newcommand{\tjSelfconsUnsupMissOneVote}{10}
\newcommand{\tjSelfconsAnyUnsup}{0.36}
\newcommand{\tjSelfconsAnyUnsupCI}{[0.22, 0.50]}
\newcommand{\tjSelfconsAnyFA}{0.02}
\newcommand{\tjSelfconsAnyPPVFive}{0.70}

\newcommand{\tjStepQUnsupMissMentionReply}{39}
\newcommand{\tjStepQUnsupMissNamesClaim}{11}
\newcommand{\tjStepQUnsupMissCapped}{6}
\newcommand{\tjStepQUnsupVoucherK}{8}
\newcommand{\tjStepQUnsupVoucherN}{8}
\newcommand{\tjStepQUnsupOtherK}{1}
\newcommand{\tjStepQUnsupOtherN}{42}

\newcommand{\tjSeqWrongToolK}{50}
\newcommand{\tjSeqWrongToolN}{50}
\newcommand{\tjSeqFA}{0}
\newcommand{\tjSeqOrganicCleanK}{27}
\newcommand{\tjSeqOrganicCleanN}{47}
\newcommand{\tjSeqLenientOrganicCleanK}{17}
\newcommand{\tjSeqOrganicRepeatEscK}{25}
\newcommand{\tjSeqOrganicNoEligK}{17}
\newcommand{\tjSeqOrganicTwoCustK}{2}
\newcommand{\tjStepLStepHitsK}{222}
\newcommand{\tjStepLStepHitsN}{297}
\newcommand{\tjStepLDistinctRationales}{15}
\newcommand{\tjAgentRefundRepliesExactK}{20}
\newcommand{\tjAgentRefundRepliesExactN}{20}
\newcommand{\tjAgentRepliesOracleIdentical}{0}
\newcommand{\tjAgentNPerStratum}{10}
\newcommand{\tjFAUpperBoundPct}{3.6\%}
\newcommand{\tjAblAFA}{0.227}

\newcommand{\tjAblAFAN}{141}
\newcommand{\tjAblAFACI}{[0.16, 0.30]}

\newcommand{\tjAblAPairReplySame}{+0.00}

\newcommand{\tjAblAPairReplyChanged}{+0.46}

\newcommand{\tjAblAPairReplyChangedCI}{[+0.37, +0.55]}
\newcommand{\tjAblAPairUnsup}{+0.14}

\newcommand{\tjAblAPairUnsupCI}{[+0.04, +0.25]}
\newcommand{\tjAblAPairPrem}{+0.78}

\newcommand{\tjAblAPairPremCI}{[+0.66, +0.89]}
\newcommand{\tjAblAPairAll}{+0.15}

\newcommand{\tjAblAPairAllCI}{[+0.12, +0.19]}

\newcommand{\tjAblBFA}{0.156}
\newcommand{\tjAblBFAK}{22}
\newcommand{\tjAblBFAN}{141}
\newcommand{\tjAblBFACI}{[0.10, 0.22]}
\newcommand{\tjAblBSameFlipK}{1}

\newcommand{\tjAblBPairReplySame}{\ensuremath{-}0.01}

\newcommand{\tjAblBPairReplySameCI}{[\ensuremath{-}0.02, +0.00]}
\newcommand{\tjAblBPairReplyChanged}{+0.49}

\newcommand{\tjAblBPairReplyChangedCI}{[+0.40, +0.58]}
\newcommand{\tjAblBPairUnsup}{+0.18}

\newcommand{\tjAblBPairUnsupCI}{[+0.06, +0.31]}
\newcommand{\tjAblBPairPrem}{+0.80}

\newcommand{\tjAblBPairPremCI}{[+0.69, +0.90]}
\newcommand{\tjAblBPairAll}{+0.16}

\newcommand{\tjAblBPairAllCI}{[+0.12, +0.19]}
\newcommand{\tjAblCFA}{0.000}

\newcommand{\tjAblCFACI}{---}

\newcommand{\tjAblCPairSameKept}{+0.50}

\newcommand{\tjAblCPairSameKeptCI}{[+0.42, +0.59]}
\newcommand{\tjAblCPairSameBroke}{+1.00}

\newcommand{\tjAblCPairReplySame}{+0.69}

\newcommand{\tjAblCPairReplySameCI}{[+0.62, +0.76]}
\newcommand{\tjAblCPairReplyChanged}{+0.60}

\newcommand{\tjAblCPairReplyChangedCI}{[+0.53, +0.67]}
\newcommand{\tjAblCPairUnsup}{+0.26}

\newcommand{\tjAblCPairUnsupCI}{[+0.14, +0.38]}
\newcommand{\tjAblCPairPrem}{+0.94}

\newcommand{\tjAblCPairPremCI}{[+0.87, +1.00]}
\newcommand{\tjAblCPairAll}{+0.66}

\newcommand{\tjAblCPairAllCI}{[+0.61, +0.72]}
\newcommand{\tjAblDFA}{0.000}

\newcommand{\tjAblDFACI}{---}

\newcommand{\tjAblDPairReplySame}{+1.00}

\newcommand{\tjAblDPairReplySameCI}{[+1.00, +1.00]}
\newcommand{\tjAblDPairReplyChanged}{+0.53}

\newcommand{\tjAblDPairReplyChangedCI}{[+0.46, +0.60]}
\newcommand{\tjAblDPairUnsup}{+0.18}

\newcommand{\tjAblDPairUnsupCI}{[+0.08, +0.29]}
\newcommand{\tjAblDPairPrem}{+0.88}

\newcommand{\tjAblDPairPremCI}{[+0.79, +0.96]}
\newcommand{\tjAblDPairAll}{+0.84}

\newcommand{\tjAblDPairAllCI}{[+0.81, +0.87]}
\newcommand{\tjAblCOne}{+0.040}
\newcommand{\tjAblCOneCI}{[\ensuremath{-}0.074, +0.159]}

\newcommand{\tjAblCOnePHolm}{0.559}
\newcommand{\tjAblCTwo}{+0.310}
\newcommand{\tjAblCTwoCI}{[+0.241, +0.382]}

\newcommand{\tjAblCTwoPHolm}{<0.001}
\newcommand{\tjAblCThree}{+0.080}
\newcommand{\tjAblCThreeCI}{[+0.000, +0.178]}

\newcommand{\tjAblCThreePHolm}{0.261}
\newcommand{\tjAblCFour}{\ensuremath{-}0.071}
\newcommand{\tjAblCFourCI}{[\ensuremath{-}0.142, +0.000]}

\newcommand{\tjAblCFourPHolm}{0.193}
\newcommand{\tjAblCFive}{+0.690}
\newcommand{\tjAblCFiveCI}{[+0.643, +0.741]}

\newcommand{\tjAblCFivePHolm}{<0.001}
\newcommand{\tjAblDAgreeN}{400}
\newcommand{\tjAblDAgreeFaultyK}{392}
\newcommand{\tjAblDAgreeTypeK}{331}
\newcommand{\tjAblDAgreeStepK}{376}
\newcommand{\tjAblDAgreeConfK}{267}
\newcommand{\tjAblDAgreeRationaleK}{0}
\newcommand{\tjAblDFlipK}{8}

\newcommand{\tjAblAAgreeN}{400}
\newcommand{\tjAblAAgreeFaultyK}{289}

\newcommand{\tjAblSchemaIdenticalN}{441}
\newcommand{\tjAblDPremStepFourK}{18}
\newcommand{\tjAblDPremStepThreeK}{6}
\newcommand{\tjAblDPremDetectedN}{44}
\newcommand{\tjAblDPremStepOtherK}{20}
\newcommand{\tjAblARecallSameMinPct}{20\%}
\newcommand{\tjAblARecallSameMaxPct}{26\%}
\newcommand{\tjAblARecallMoveMaxPts}{54}
\newcommand{\tjAblAFAOrigK}{23}
\newcommand{\tjAblAFAOrigN}{100}
\newcommand{\tjAblCRefundStepK}{100}
\newcommand{\tjAblCRefundStepN}{100}

\newcommand{\tjAblCSkippedSilentN}{25}
\newcommand{\tjAblCHallucK}{32}
\newcommand{\tjAblCWrongToolK}{6}
\newcommand{\tjOrgDMissK}{7}

\newcommand{\tjOrgAFlagsFaultyK}{11}
\newcommand{\tjOrgAFlagsFaultyN}{13}
\newcommand{\tjOrgAFlagsCleanK}{18}
\newcommand{\tjOrgAFlagsCleanN}{47}
\newcommand{\tjOrgDFlagsFaultyK}{6}
\newcommand{\tjOrgDFlagsFaultyN}{13}
\newcommand{\tjOrgDFlagsCleanK}{4}

\newcommand{\tjPairExOrder}{ORD-20000}
\newcommand{\tjPairExAmount}{398.50}

\long\edef\tjDashes{\string-\string-\string-}
\newcommand{\tjCI}[1]{\ifx#1\tjDashes\else\ci{#1}\fi}
\def\tjPrel#1#2\relax{\ifx<#1\else=\fi#1#2}
\newcommand{\tjPval}[1]{\ifx#1\tjDashes\else, $p$\,\expandafter\tjPrel#1\relax\fi}
\newcommand{\tjOrNA}[1]{\ifx#1\tjDashes n/a\else#1\fi}

\newif\ifablation
\ablationtrue

\newif\ifreleased
\releasedtrue

\newif\ifchanges
\changesfalse

\definecolor{cigray}{gray}{0.45}
\newcommand{\ci}[1]{{\scriptsize\textcolor{cigray}{#1}}}
\newcommand{\ftype}[1]{\texttt{#1}}
\newcommand{\judge}[1]{\texttt{#1}}

\title{trajectory-judge: What Outcome-Only LLM Judges\\Miss on Agent Trajectories}

\author{%
  Hadi Mohammadi\\
  Department of Methodology, Statistics \& Data Science\\
  Utrecht University, The Netherlands\\
  \texttt{h.mohammadi@uu.nl}\\[1.6ex]
  {\normalfont\small
  \href{https://github.com/mohammadi-hadi/trajectory-judge}{\faGithub\enspace\texttt{github.com/mohammadi-hadi/trajectory-judge}}}\\[0.3ex]
  {\normalfont\small
  \href{https://pypi.org/project/trajectory-judge/}{\faPython\enspace\texttt{pip install trajectory-judge\ifreleased==0.2.0\fi}}\qquad
  \href{https://doi.org/10.5281/zenodo.21797926}{\faArchive\enspace\texttt{doi:10.5281/zenodo.21797926}}}
}

\begin{document}

\maketitle

\begin{abstract}
A direct test of an LLM judge of agent trajectories injects faults into correct runs and reports
recall, per fault type or by whether the fault broke the environment outcome (\emph{loud}) or not
(\emph{silent}). Such recall can credit a judge with detection it does not have; \emph{paired
discrimination}, its flag rate on the faults minus its rate on the clean runs they came from,
exposes this. Our testbed, a deterministic support desk with a scripted oracle and a one-step
fault injector, labels all \tjNTraj{} trajectories exactly. A 14B judge shown only the request
and final reply scores \tjOutcomeReplySameRecallMinPct{} to \tjOutcomeReplySameRecallMaxPct{}
recall on four fault types that leave the reply unchanged. There its input is the clean run's,
so its paired discrimination is zero and that recall is its flag rate on clean runs. Splitting by outcome
survival does not fix this: its loud recall of \tjOutcomeLoudPct{} is a paired
\tjOutcomePairLoud{} and its silent recall of \tjOutcomeSilentPct{} a paired
\tjOutcomePairSilent, all from the two fault types that change the reply. Told to check each
step, the same model flags every fault of those four types and \tjStepQFAK{} of \tjNClean{} clean
runs (95\% CI up to \tjFAUpperBoundPct). It does not reliably check the reply: of four
invented promises it flags one every time and the other three once in \tjStepQUnsupOtherN{}
faults.
\ifablation
Shown every step but asked only about the reply, it still reaches a paired
\tjAblCPairReplySame{} on reply-unchanged faults, against \tjAblDPairReplySame{} when told to
check each step.
\fi
We recommend reporting paired discrimination against clean parents, split by whether the fault
reaches the judge's input and by outcome survival, and release the testbed, raw verdicts and
analysis pipeline.
\end{abstract}

\section{Introduction}
\label{sec:intro}

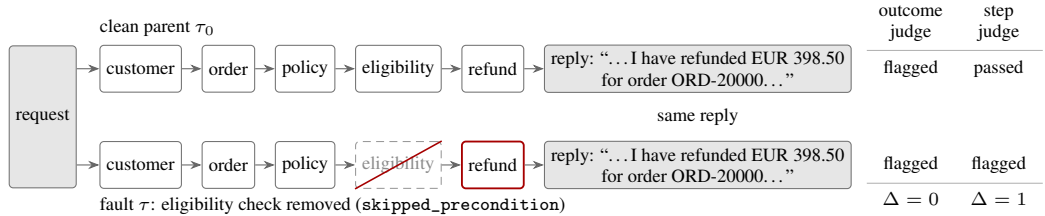
\begin{figure}[t]
\centering
\begin{tikzpicture}[
  font=\scriptsize,
  node distance=2.6mm,
  st/.style={draw=black!55, rounded corners=1.5pt, align=center, inner sep=2.5pt,
             minimum height=6.2mm},
  seen/.style={fill=black!10},
  gone/.style={st, draw=black!40, densely dashed, text=black!45},
  bad/.style={st, draw=red!65!black, line width=0.7pt},
  vd/.style={align=center, inner sep=1pt, minimum width=11.5mm},
  arr/.style={-{Stealth[length=1.5mm]}, black!60}
]
\node[st] (a0) {customer};
\node[st, right=of a0] (a1) {order};
\node[st, right=of a1] (a2) {policy};
\node[st, right=of a2] (a3) {eligibility};
\node[st, right=of a3] (a4) {refund};
\node[st, seen, right=of a4] (a5)
  {reply: ``\ldots I have refunded EUR \tjPairExAmount\\for order \tjPairExOrder\ldots''};
\node[st, below=6.4mm of a0] (b0) {customer};
\node[st, right=of b0] (b1) {order};
\node[st, right=of b1] (b2) {policy};
\node[gone] (b3) at (a3 |- b0) {eligibility};
\node[bad] (b4) at (a4 |- b0) {refund};
\node[st, seen] (b5) at (a5 |- b0)
  {reply: ``\ldots I have refunded EUR \tjPairExAmount\\for order \tjPairExOrder\ldots''};
\draw[red!65!black, line width=0.6pt] (b3.south west) -- (b3.north east);
\node[st, seen, left=3mm of $(a0.west)!0.5!(b0.west)$, minimum height=18.8mm] (g) {request};
\foreach \r in {a,b} {
  \draw[arr] (g.east |- \r0) -- (\r0);
  \draw[arr] (\r0) -- (\r1); \draw[arr] (\r1) -- (\r2);
  \draw[arr] (\r2.east) -- (\r3.west); \draw[arr] (\r3.east) -- (\r4.west);
  \draw[arr] (\r4) -- (\r5);
}
\node[font=\scriptsize, inner sep=0pt] at ($(a5.south)!0.5!(b5.north)$) {same reply};
\node[anchor=south west, inner sep=0pt] at ($(a0.north west)+(0,1.1mm)$)
  {clean parent $\tau_0$};
\node[anchor=north west, inner sep=0pt] at ($(b0.south west)+(0,-1.1mm)$)
  {fault $\tau$: eligibility check removed (\ftype{skipped\_precondition})};
\node[vd, right=2mm of a5] (v1a) {flagged};
\node[vd, right=0mm of v1a] (v2a) {passed};
\node[vd] (v1b) at (v1a |- b0) {flagged};
\node[vd] (v2b) at (v2a |- b0) {flagged};
\node[vd, anchor=south] at ($(v1a.north)+(0,1.7mm)$) {outcome\\judge};
\node[vd, anchor=south] at ($(v2a.north)+(0,1.7mm)$) {step\\judge};
\node[vd, anchor=north] at ($(v1b.south)+(0,-1.7mm)$) {$\Delta=0$};
\node[vd, anchor=north] at ($(v2b.south)+(0,-1.7mm)$) {$\Delta=1$};
\draw[black!30] ($(v1a.north west)+(0,0.9mm)$) -- ($(v2a.north east)+(0,0.9mm)$);
\draw[black!30] ($(v1b.south west)+(0,-0.9mm)$) -- ($(v2b.south east)+(0,-0.9mm)$);
\end{tikzpicture}
\caption{A fault and its clean parent. Shaded: what the outcome judge reads.
The reply does not change, so it flags both runs; the step judge flags only the fault.}
\label{fig:pair}
\end{figure}

Agents that call tools are often gated on outcome-level judgment: a model reads the user's
request and the agent's final reply and decides whether the case was handled well. Some faults
never reach the reply. A refund of the correct amount issued without the required eligibility
check produces the same reply as one issued correctly (Figure~\ref{fig:pair}). A direct test injects faults into correct
runs and reports recall, per fault type or by whether the fault broke the environment outcome
(loud) or not (silent). We show that such recall
can credit a judge with detection it does not have, and that pairing each injected fault with
the clean run it came from, while labelling whether the fault reaches each judge's input,
exposes this.

That a judge shown only the request and the reply cannot see a fault absent from the reply is
expected, and serves here as a control with a known answer. What pairing measures is how far
recall overstates detection. Recall inherits the judge's base rate in each scenario: our injector
leaves the reply byte-identical to the clean run's on four of six fault types, and a 14B judge
shown only the reply scores \tjOutcomeReplySameRecallMinPct{} to
\tjOutcomeReplySameRecallMaxPct{} recall on them, yet its paired discrimination $\Delta$ there (flag rate
on the faults minus flag rate on their clean parents) is zero. That recall is its flag rate on
clean runs, \tjOutcomeFAStratumMinPct{} to \tjOutcomeFAStratumMaxPct{} by scenario\ifablation;
on a newer inference engine the same prompt scores \tjAblARecallSameMinPct{} to
\tjAblARecallSameMaxPct{} there, again with no discrimination (\S\ref{sec:viewtask})\fi.
Splitting recall by outcome survival is not sufficient. Of the judge's \tjOutcomeLoudFlags{}
flags on loud faults, \tjOutcomeLoudFlagsSame{} are reply-unchanged faults flagged as often as
their parents, and after pairing its loud recall of \tjOutcomeLoudPct{} is \tjOutcomePairLoud,
all of it from the one loud type that changes the reply. The reply itself is discussed but rarely
checked: told to check each step, the same model's paired discrimination on an invented promise
(\tjStepQPairUnsupClaim) is close to the reply-only judge's (\tjOutcomePairUnsupClaim), mostly from
one of the four invented sentences.
\ifablation
The two judges also differ in instruction; an ablation crosses it with view
(\S\ref{sec:viewtask}).
\fi

Evaluating an evaluator usually needs human labels, which puts the labeller under test too. We
avoid this by construction: a deterministic support-desk environment, a scripted
oracle that is correct by construction and checked by a test on every generated instance, and a
fault injector that makes one edit to an oracle run at a known step and replays the edited
calls, so every observation comes from the environment. Each trajectory carries an exact label at no annotation
cost: faulty or not, step, type, whether the environment outcome survived, whether the reply
changed, and its clean parent. With the agent fixed, an error belongs to the judge or to its
view, not to the agent or to annotators.

\textbf{Contributions.}
(i)~A measurement design for trajectory judges: each injected fault is paired with its clean
parent, the judge's own control, and labelled by whether it reaches each judge's input and
whether the environment outcome survived (\S\ref{sec:testbed}, \S\ref{sec:metrics}).
(ii)~A testbed that implements it: a permissive environment with a strict checker, a
correct-by-construction oracle, a six-type single-fault injector, and a rule baseline whose
coverage is pinned by tests (\S\ref{sec:testbed}).
(iii)~Findings scoped to this testbed and two local models: on reply-unchanged faults a
reply-only judge's recall is its base rate; outcome-stratified recall needs pairing; a judge told
to check each step flags every reply-unchanged fault and \tjStepQFAK{} of \tjNClean{} clean runs;
no judge reliably verifies the reply; majority voting over three samples adds cost without a
measurable gain; an 8B judge does not discriminate under our prompt (\S\ref{sec:results},
\S\ref{sec:analysis}).
(iv)~An artifact that rebuilds every number in this paper offline from the released raw
verdicts (\S\ref{sec:repro}).

\section{Related work}
\label{sec:related}

\textbf{Reliability of LLM judges.} Model-as-judge evaluation is standard for open-ended
outputs \citep{zheng2023judging,liu2023geval,gu2024survey}, and its pathologies are
well documented: position bias \citep{wang2023fair}, verbosity bias \citep{zheng2023judging},
self-preference \citep{panickssery2024self}, objective failures on instruction-following
comparisons \citep{zeng2024llmbar}, and leniency and imperfect agreement with human scores even
on a simple task \citep{thakur2024judging}. That line audits judges of single responses against human
labels; we audit judges of multi-step tool trajectories against labels correct by
construction, so a flag on a clean trajectory is a false positive.
The judge design itself (reasoning before the verdict) follows \citet{mohammadi2026evalmoraal},
and treating a model's labels as measurements that need their own reliability estimate follows
\citet{mohammadi2025assessing}; \citet{mohammadi2026let} develops both at thesis length and notes
that a model's stated reasoning may not reflect how it decided
\citep[see also][]{turpin2023language,mohammadi2025faithful,mohammadi2025explainability}.

\textbf{Agent benchmarks.} Benchmarks for tool-using agents score the agent, increasingly on
the trajectory rather than the outcome alone: $\tau$-bench and its successor compare final
database state \citep{yao2024tau,barres2026tau2}, AgentBench and WebArena score task success \citep{liu2024agentbench,
zhou2024webarena}, AgentBoard tracks subgoal progress \citep{ma2024agentboard}, and recent
work scores reasoning and tool-use trajectories directly \citep{kim2025beyond,he2025trajectbench}. We
invert the roles: the agent is a fixed oracle, and the evaluator is the system under test.

\textbf{Monitoring and error detection.} AI-control monitors score actions or transcripts for
covert sabotage \citep{bhatt2025ctrlz,kutasov2025shade}, and step-level guardrails flag unsafe
tool calls \citep[TS-Bench;][]{mou2026toolsafe}. Judges of agent trajectories are checked
against benchmark outcomes or human labels
\citep{pan2024autonomous,lu2025agentrewardbench,fan2026agentprocessbench}, and an agentic judge inspecting
intermediate work outperforms an LLM judge \citep{zhuge2025agent}. What the evaluator sees
matters: chain-of-thought monitors can far outperform monitors of actions and outputs
\citep{baker2025monitoring}, state-based success checks credit $\tau$-bench runs that violate
the procedure or return nothing \citep{cao2026corrupt,zhu2025establishing}, and LLM judges struggle to catch false completion claims
\citep{advani2026false}.

\textbf{Failure attribution and fault injection.} Locating the failing step or agent in a
trace is an emerging task \citep{zhang2025whowhen,deshpande2025trail,cemri2025mast,ma2025failure}.
Injected faults supply labels for failure attribution \citep{zhang2025agentracer} and for
judges of dialogue compliance \citep{yang2026complibench} and of research-agent traces
\citep{wang2026reflect}. Paired perturbations expose evaluator
blind spots on single responses \citep{doddapaneni2024finding}, and concurrent work tests
localisation on view-identical trace pairs \citep{zhu2026telemetrysuffbench}. Concurrent work
also scores judges on matched fault and clean twins, by catch minus false rejection, for
pre-execution oversight \citep{han2026rejective}. What we add is a label on each fault for whether
it reaches each judge's input, the propagation condition that separates weak from strong mutation
\citep{howden1982weak}, and what follows from it for post-hoc trajectory judges. Without
the pairing, per-type and outcome-stratified recall can credit a judge with detection it lacks
(\S\ref{sec:blindspot}). Mutation testing has seeded known defects to measure a detector for half
a century \citep{demillo1978hints,jia2011analysis}. Whether mutants substitute for real
faults is the classic validity question \citep{just2014mutants}; our paired results cover injected
faults, whose mix is unlike an agent's own (\S\ref{sec:organic}).

\textbf{Process versus outcome supervision.} In mathematics, step-level feedback trains better
verifiers than outcome-level feedback \citep{lightman2024verify,wang2024mathshepherd} and is
needed for correct reasoning steps even when final-answer error is similar
\citep{uesato2022solving}; ProcessBench measures error localisation in reasoning chains
\citep{zheng2024processbench}, on which recent process reward models are evaluated
\citep{zhang2025lessons}. That agenda trains scorers where the final answer is checkable; we ask the measurement question
underneath it, whether an off-the-shelf judge can see process faults at all when they
leave the final reply unchanged \citep[cf.][]{lang2024deceive}, and find that one reading only
the request and reply cannot, while the same model told to check each step can.

\textbf{Calibration.} We score each judge's confidence with expected calibration error
\citep{naeini2015obtaining,guo2017calibration} and the Brier score. Confidence read from token probabilities
\citep[e.g.][]{kadavath2022language,mohammadi2026reliability} or verbalised \citep{tian2023just} can be usable, but verbalised
confidence trends overconfident \citep{xiong2024can}.

\section{A testbed with ground truth by construction}
\label{sec:testbed}

\textbf{Environment.} The environment is a customer-support desk with seven tools
(\texttt{get\_customer}, \texttt{lookup\_order}, \texttt{get\_policy},
\texttt{check\_eligibility}, \texttt{issue\_refund}, \texttt{escalate}, \texttt{reply}) and a
written standard operating procedure: verify the customer, look up the order, read the policy
for the item, confirm eligibility before moving money, refund exactly the authorised amount,
escalate when not eligible, and reply asserting only what the observations support. Instances
are generated in six scenarios: full-price refunds, restocking fees, expired windows,
non-refundable items, orders belonging to a different customer, and orders already
refunded, drawn round-robin from a single seeded stream (Appendix~\ref{app:env}). The
environment is permissive and the checker is strict: \texttt{issue\_refund} will
refund an order whose eligibility was never checked, exactly as a real payments API would.
Nothing in the world stops an agent from skipping the process; only the rule checker says it
was wrong. Without this there would be no silent faults to measure, so a test
(\texttt{test\_environment\_is\_permissive\_by\_design}) pins it.

\textbf{Notation.} A trajectory $\tau=(g, s_0,\dots,s_{T-1}, a)$ pairs a goal $g$ and final
reply $a$ with steps $s_t=(h_t, c_t, o_t)$: a thought, a tool call
$c_t=(\mathrm{tool}_t,\mathrm{args}_t)$, and an observation $o_t$. Its label
$\ell(\tau)=(y, t^{*}, \phi, \omega)$ records faulty or not, the failure step, the type
$\phi\in\Phi$ ($|\Phi|=6$), and the \emph{environment outcome} $\omega$: whether the episode ends
in the refund amount or escalation the instance requires, with reply text excluded. A fault is
\emph{silent} iff $y\wedge\omega$ and \emph{loud} iff $y\wedge\neg\omega$, and every faulty
$\tau$ has a \emph{clean parent} $\tau_0$, the oracle run it was injected into. A judge maps a
\emph{view} of $\tau$ to a verdict $(\hat y, \hat t, \hat\phi, \hat c, r)$ with stated
confidence $\hat c$ and rationale $r$. The step view $V_{\mathrm{step}}(\tau)$ is the full
rendering. The outcome-only view $V_{\mathrm{out}}(\tau)=(g,a)$ follows LLM-as-judge usage
\citep{zheng2023judging} and holds the request and the reply, never the end state. We call the
judge on this view the \emph{outcome judge}. Unlike a state check in the style of $\tau$-bench
\citep{yao2024tau}, which flags every loud fault, it cannot see a loud fault that the reply does
not reveal. Every fault changes
$V_{\mathrm{step}}$, and $V_{\mathrm{out}}(\tau)=V_{\mathrm{out}}(\tau_0)$ iff the fault leaves
the reply unchanged. For any judge whose verdict depends only on $(g,a)$, such a fault has its
parent's verdict distribution, so its expected paired discrimination (\S\ref{sec:metrics}) is
zero; under greedy decoding in the August runs (\S\ref{sec:judges}) we observe identical verdicts, confidences and
rationales on \tjOutcomeIdenticalPairs{} of \tjNPairsReplySame{} pairs. This does not cover judges given tools,
retrieval or the environment state, and the share of faults it covers (\tjNReplySame{} of
\tjNFaulty) is set by our injector.

\textbf{Oracle.} The oracle is a fixed six-step script, identical across scenarios up to one
branch: verify, look up, read the policy for the SKU returned by the lookup
observation, check eligibility, then refund the authorised amount if eligible and escalate
otherwise, and reply. A test parametrised over all instances asserts it violates no rule and
reaches the expected outcome.

\begin{table}[t]
\centering
\caption{The six fault types, \tjNPerType{} faults each.}
\label{tab:taxonomy}
\small
\setlength{\tabcolsep}{4pt}
\begin{tabular}{@{}llccc@{}}
\toprule
Fault type & The single edit & \begin{tabular}[b]{@{}c@{}}Rule\\ recall\end{tabular} & \begin{tabular}[b]{@{}c@{}}Env.\ outcome\\ survives\end{tabular} & \begin{tabular}[b]{@{}c@{}}Reply\\ changed\end{tabular} \\
\midrule
\ftype{wrong\_tool} & order fetched again in place of the policy & \tjRulesWrongTool & always & no \\
\ftype{hallucinated\_argument} & policy fetched for an invented SKU & \tjRulesHallucArg & always & no \\
\ftype{skipped\_precondition} & refund without the eligibility check & \tjRulesSkippedPre & at full price & no \\
\ftype{ignored\_observation} & refund of an unauthorised amount & \tjRulesIgnoredObs & never & no \\
\ftype{premature\_stop} & stops before acting or replying & \tjRulesPrematureStop & never & yes \\
\ftype{unsupported\_claim} & invented promise added to the reply & \tjRulesUnsupClaim & always & yes \\
\bottomrule
\end{tabular}
\end{table}

\textbf{Fault injector.} Each mutation edits the oracle's call list at one step and
replays the edited list against a fresh environment, so a hallucinated SKU fails its
lookup and no observation is written by hand; a test asserts that the replayed observations are
consistent with the calls. Each mutation targets one failure: \ftype{skipped\_precondition},
for example, refunds the order total so that the amount stays grounded in the lookup and no
second checker rule fires. Four types leave the reply unchanged (\tjNReplySame{} faults: \tjNReplySameKept{}
silent, \tjNReplySameBroke{} loud); \ftype{premature\_stop} replaces it with a holding message
and \ftype{unsupported\_claim} appends a sentence (Table~\ref{tab:taxonomy}). Keeping the
oracle's reply has one side effect a real run would not have. In the \tjNReplySameBroke{} loud
faults (all of \ftype{ignored\_observation}, and \ftype{skipped\_precondition} under a
restocking fee) the reply states the authorised amount although the order total was refunded.
The agent of \S\ref{sec:organic} states the amount it refunded in
\tjAgentRefundRepliesExactK{} of \tjAgentRefundRepliesExactN{} refund replies, and
\tjAgentRepliesOracleIdentical{} of its \tjAgentN{} replies equal the oracle's. The stale amount
is invisible to a judge that reads only the reply, but it gives one that reads the steps a
second cue. Injection is string-seeded per (instance, type); the judged set is \tjNTraj{}
trajectories (\tjNClean{} clean, \tjNSilent{} silent, \tjNLoud{} loud) in a fixed shuffled order.

\textbf{The rule engine and its two zeros.} A checker walks each trajectory once, keeping a
pool of grounded values seeded from the goal, and enforces eight rules over grounding, ordering,
amounts, identity, and terminality (Appendix~\ref{app:checker}). It catches every fault of four
types and none of the other two. These zeros are limits of the eight rules: none encodes the
required tool order, so a grounded but redundant \texttt{lookup\_order} in place of
\texttt{get\_policy} breaks nothing, and none reads the reply, since prose is excluded from the
grounding pool. A required-sequence rule, written post hoc and kept out of the baseline, catches
\tjSeqWrongToolK{} of \tjSeqWrongToolN{} \ftype{wrong\_tool} faults with \tjSeqFA{} false alarms
on clean runs, but also flags \tjSeqOrganicCleanK{} of the \tjSeqOrganicCleanN{} agent episodes
the checker passes (Appendix~\ref{app:checker}). The rule engine is thus a fixed reference point
that does not bound programmatic checking; a test fails if either zero changes.

\section{Five judges}
\label{sec:judges}

\begin{table}[t]
\centering
\caption{The five judges (prompts and schemas: Appendix~\ref{app:prompts}).}
\label{tab:judges}
\small
\setlength{\tabcolsep}{4pt}
\begin{tabular}{@{}llllll@{}}
\toprule
Judge & View & Task & Model & Decoding & Confidence \\
\midrule
\judge{rules} & steps & eight rules & n/a & deterministic & two constants \\
\judge{outcome} & $(g,a)$ & judge the reply & \texttt{qwen2.5:14b} & $T{=}0$, seed 7 & stated \\
\judge{step 14B} & full & check each step & \texttt{qwen2.5:14b} & $T{=}0$, seed 7 & stated \\
\judge{step 8B} & full & check each step & \texttt{llama3.1:8b} & $T{=}0$, seed 7 & stated \\
\judge{selfcons} & full & check each step & \texttt{qwen2.5:14b} & $T{=}0.7$, seeds 7--9 & vote share \\
\bottomrule
\end{tabular}
\end{table}

Table~\ref{tab:judges} lists the five judges \citep{qwen2025qwen,dubey2024llama}. The LLM judges
share the procedure and taxonomy word for word and decode under a JSON schema with
\texttt{reasoning} first. The outcome and step judges differ in view, task instruction and schema
(the outcome judge names no step and is told that unseen steps are not evidence of a failure), so
comparing them compares configurations; on reply-unchanged faults the view is necessary whatever
the instruction (\S\ref{sec:testbed})\ifablation, and \S\ref{sec:viewtask} crosses the two\fi.
The ensemble \citep{wang2023selfconsistency} is a majority over three step-judge samples at
$T{=}0.7$; its confidence, the vote share, takes two values. The LLM judges ran in August 2026 on
Ollama 0.30.11 (the August runs; code release \texttt{v0.1.0})\ifablation; the ablation of
\S\ref{sec:viewtask} ran in October on 0.33.3\fi. Prompts are verbatim in
Appendix~\ref{app:prompts}.

\section{Metrics and uncertainty}
\label{sec:metrics}

Detection is binary per trajectory. We lead with sensitivity (recall over all, silent
and loud faults) and specificity (one minus the false-alarm rate on clean runs), which, unlike
precision, F1, Brier score and ECE, do not depend on the share of faults in the set (here
\tjNFaulty{} of \tjNTraj; Appendix~\ref{app:full}).
Paired discrimination $\Delta_J$ is the mean of
$\hat y_J(\tau)-\hat y_J(\tau_0)$ over faults $\tau$ whose parent was judged (\tjNPairs{} of
\tjNFaulty): the share of faults judge $J$ flags minus the share of their clean parents it
flags. Recall mixes detection with the judge's flag rate in the fault's scenario; $\Delta$
removes the latter\ifablation. The ablation (\S\ref{sec:viewtask}) also judges the other
\tjNLateParents{} parents\fi.
Localisation is exact match on the failure step, over the faults a judge detected (a
missing or out-of-range step is a miss) and jointly over all \tjNFaulty{} faults (detected and
exact). \ftype{premature\_stop} is labelled at its last executed step, so a judge that names
the missing next action scores a miss (Appendix~\ref{app:injector}); we also report the other
five types alone (Table~\ref{tab:extra}).
Typing is six-class macro-F1 over faulty trajectories and the share of all faults
detected and correctly typed; calibration is ten-bin ECE \citep{naeini2015obtaining}
against the binary verdict, coarse for the rule engine's two constants and the ensemble's
two-valued vote share; cost is seconds per trajectory on one workstation (Brier scores
and tokens: Appendix~\ref{app:full}).

Intervals are 95\% percentile bootstrap intervals
\citep{efron1979bootstrap} ($B{=}10{,}000$) that resample trajectories within the eight design
cells (clean, and each fault type by outcome), one resample shared across judges. The rule
engine is constant within a cell, so its entries get no such interval. Pooled paired estimates
resample instances, since one parent serves several faults. Recalls and false-alarm rates at $0$ or $1$ in Table~\ref{tab:main} get Clopper--Pearson intervals ($\dagger$), and
single-type paired rows get exact McNemar tests. Intervals are conditional on the generation
seed and greedy decoding (Appendix~\ref{app:bootstrap}).

\section{Results}
\label{sec:results}

\begin{table}[t]
\centering
\caption{Main results on \tjNTraj{} trajectories. Brackets: 95\% intervals; $\dagger$: exact Clopper--Pearson.}
\label{tab:main}
\small
\setlength{\tabcolsep}{3.1pt}
\begin{tabular}{@{}lccccccccc@{}}
\toprule
 & \multicolumn{3}{c}{Recall} & False & Paired & \multicolumn{2}{c}{Localisation} & Type & \\
\cmidrule(lr){2-4}\cmidrule(lr){7-8}
Judge & all & silent & loud & alarms & $\Delta$ & detected & joint & F1 & ECE \\
\midrule
\judge{rules} & \tjRulesSens & \tjRulesSilent & \tjRulesLoud & \tjRulesFA & $\tjRulesPairAll$ & \tjRulesLocDet & \tjRulesLocJoint & \tjRulesTypeF & \tjRulesECE \\[-1pt]
 & & & \ci{\tjRulesLoudCI$^\dagger$} & \ci{\tjRulesFACI$^\dagger$} & \ci{\tjRulesPairAllCI} & & & & \\[2pt]
\judge{outcome} & \tjOutcomeSens & \tjOutcomeSilent & \tjOutcomeLoud & \tjOutcomeFA & $\tjOutcomePairAll$ & n/a & n/a & \tjOutcomeTypeF & \tjOutcomeECE \\[-1pt]
 & \ci{\tjOutcomeSensCI} & \ci{\tjOutcomeSilentCI} & \ci{\tjOutcomeLoudCI} & \ci{\tjOutcomeFACI} & \ci{\tjOutcomePairAllCI} & & & \ci{\tjOutcomeTypeFCI} & \\[2pt]
\judge{step 14B} & \tjStepQSens & \tjStepQSilent & \tjStepQLoud & \tjStepQFA & $\tjStepQPairAll$ & \tjStepQLocDet & \tjStepQLocJoint & \tjStepQTypeF & \tjStepQECE \\[-1pt]
 & \ci{\tjStepQSensCI} & \ci{\tjStepQSilentCI} & \ci{\tjStepQLoudCI} & \ci{\tjStepQFACI$^\dagger$} & \ci{\tjStepQPairAllCI} & \ci{\tjStepQLocDetCI} & \ci{\tjStepQLocJointCI} & \ci{\tjStepQTypeFCI} & \\[2pt]
\judge{step 8B} & \tjStepLSens & \tjStepLSilent & \tjStepLLoud & \tjStepLFA & $\tjStepLPairAll$ & \tjStepLLocDet & \tjStepLLocJoint & \tjStepLTypeF & \tjStepLECE \\[-1pt]
 & \ci{\tjStepLSensCI} & \ci{\tjStepLSilentCI} & \ci{\tjStepLLoudCI$^\dagger$} & \ci{\tjStepLFACI$^\dagger$} & \ci{\tjStepLPairAllCI} & \ci{\tjStepLLocDetCI} & \ci{\tjStepLLocJointCI} & \ci{\tjStepLTypeFCI} & \\[2pt]
\judge{selfcons} & \tjSelfconsSens & \tjSelfconsSilent & \tjSelfconsLoud & \tjSelfconsFA & $\tjSelfconsPairAll$ & \tjSelfconsLocDet & \tjSelfconsLocJoint & \tjSelfconsTypeF & \tjSelfconsECE \\[-1pt]
 & \ci{\tjSelfconsSensCI} & \ci{\tjSelfconsSilentCI} & \ci{\tjSelfconsLoudCI} & \ci{\tjSelfconsFACI} & \ci{\tjSelfconsPairAllCI} & \ci{\tjSelfconsLocDetCI} & \ci{\tjSelfconsLocJointCI} & \ci{\tjSelfconsTypeFCI} & \\
\bottomrule
\end{tabular}
\end{table}

\begin{figure}[t]
\centering
\includegraphics[width=\linewidth]{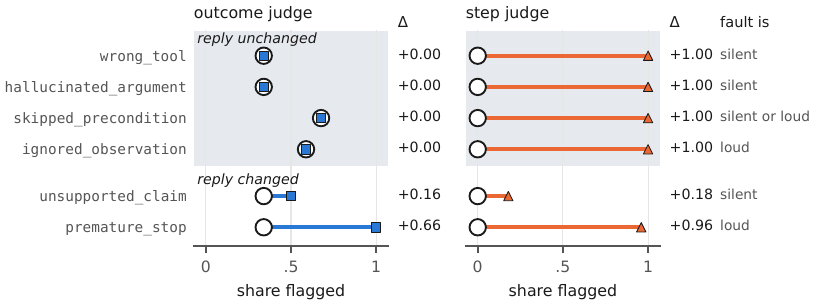}
\caption{Share of clean parents (ring) and of their faults (filled) that each single-pass 14B judge flags, by fault type; $\Delta$ is the difference. Shaded: types that leave the reply unchanged.}
\label{fig:bytype}
\end{figure}

Table~\ref{tab:main} gives the comparison; Figure~\ref{fig:bytype} sets the two single-pass 14B
judges' flag rate on each fault type against their rate on its clean parents.
The outcome judge flags \tjOutcomeLoud{} \ci{\tjOutcomeLoudCI} of loud faults and
\tjOutcomeSilent{} \ci{\tjOutcomeSilentCI} of silent ones, and \tjOutcomeFA{}
\ci{\tjOutcomeFACI} of clean runs: much of that recall is not detection, and its paired $\Delta$
is $\tjOutcomePairAll$ (\S\ref{sec:blindspot}). The step judge, the same model told to check every step, reaches
\tjStepQSilent{} \ci{\tjStepQSilentCI} silent recall with \tjStepQFAK{} false alarms in
\tjNClean{} clean runs (95\% CI up to \tjFAUpperBoundPct). That is too few clean runs to certify a
production gate: if faults were 5\% of traffic, its precision would be \tjStepQPPVFive{} at the
observed rates but \tjStepQPPVFiveLow{} at the top of that interval, and the outcome judge's
\tjOutcomePPVFive{} \ci{\tjOutcomePPVFiveCI}. The step judge localises every fault it detects on
the five types whose fault is an executed step (\tjStepQLocNonOmitK{} of \tjStepQLocNonOmitN). On
\ftype{premature\_stop}, whose fault is an action never taken, \tjStepQLocInvalid{} verdicts give
a step outside the trajectory; counted as misses, they bring its localisation to \tjStepQLocDet{} of
detected faults and \tjStepQLocJoint{} of all faults. Its ECE at this set's fault prevalence is
\tjStepQECE, the outcome judge's \tjOutcomeECE{} (Figure~\ref{fig:calibration} in
Appendix~\ref{app:full}). The rule
engine costs nothing, raises no false alarm on oracle runs (they satisfy every rule by
construction) and types faults best (\tjRulesTypeF{} macro-F1) because it never guesses. It
misses more than half of the silent faults, since two of the six types fall outside its eight
rules (\S\ref{sec:testbed}).

\section{Analysis}
\label{sec:analysis}

The consequences drawn below follow from this benchmark; none was tested in deployment.

\subsection{Recall is not detection}
\label{sec:blindspot}

\begin{table}[t]
\centering
\caption{Recall and, in parentheses, paired $\Delta$, by whether the fault changes the reply and whether it is silent or loud.}
\label{tab:paired}
\footnotesize
\setlength{\tabcolsep}{3.5pt}
\begin{tabular}{@{}llrccccc@{}}
\toprule
Reply & Fault & $n$ (pairs) & \judge{rules} & \judge{outcome} & \judge{step 14B} & \judge{step 8B} & \judge{selfcons} \\
\midrule
unchanged & silent & \tjNReplySameKept{} (\tjNPairsReplySameKept) & \tjRulesReplySameKept{} ($\tjRulesPairReplySameKept$) & \tjOutcomeReplySameKept{} ($\tjOutcomePairReplySameKept$) & \tjStepQReplySameKept{} ($\tjStepQPairReplySameKept$) & \tjStepLReplySameKept{} ($\tjStepLPairReplySameKept$) & \tjSelfconsReplySameKept{} ($\tjSelfconsPairReplySameKept$) \\
unchanged & loud & \tjNReplySameBroke{} (\tjNPairsReplySameBroke) & \tjRulesReplySameBroke{} ($\tjRulesPairReplySameBroke$) & \tjOutcomeReplySameBroke{} ($\tjOutcomePairReplySameBroke$) & \tjStepQReplySameBroke{} ($\tjStepQPairReplySameBroke$) & \tjStepLReplySameBroke{} ($\tjStepLPairReplySameBroke$) & \tjSelfconsReplySameBroke{} ($\tjSelfconsPairReplySameBroke$) \\
changed & silent & \tjNPerType{} (\tjPairNUnsupClaim) & \tjRulesUnsupClaim{} ($\tjRulesPairUnsupClaim$) & \tjOutcomeUnsupClaim{} ($\tjOutcomePairUnsupClaim$) & \tjStepQUnsupClaim{} ($\tjStepQPairUnsupClaim$) & \tjStepLUnsupClaim{} ($\tjStepLPairUnsupClaim$) & \tjSelfconsUnsupClaim{} ($\tjSelfconsPairUnsupClaim$) \\
changed & loud & \tjNPerType{} (\tjPairNPrematureStop) & \tjRulesPrematureStop{} ($\tjRulesPairPrematureStop$) & \tjOutcomePrematureStop{} ($\tjOutcomePairPrematureStop$) & \tjStepQPrematureStop{} ($\tjStepQPairPrematureStop$) & \tjStepLPrematureStop{} ($\tjStepLPairPrematureStop$) & \tjSelfconsPrematureStop{} ($\tjSelfconsPairPrematureStop$) \\
\midrule
\multicolumn{2}{@{}l}{clean runs: false alarms} & \tjNClean & \tjRulesFATwo & \tjOutcomeFATwo & \tjStepQFATwo & \tjStepLFATwo & \tjSelfconsFATwo \\
\bottomrule
\end{tabular}
\end{table}

Recall counts how often a judge flags faulty runs. Whether it would have flagged the same runs
without the fault is a separate question, and each fault's clean parent answers it:
Table~\ref{tab:paired} reports paired $\Delta$ (\S\ref{sec:metrics}) beside recall.

On the \tjNReplySame{} faults of the four reply-unchanged types the outcome judge receives its
parent's prompt, so its expected $\Delta$ is zero whatever the model (\S\ref{sec:testbed}). Its
recall on these types runs from \tjOutcomeReplySameRecallMinPct{} to
\tjOutcomeReplySameRecallMaxPct{} with $\Delta = 0$ on each. A fault is flagged when its clean reply is, and the judge's rate on clean replies depends
on the scenario: \tjOutcomeFAHappyK{} of \tjOutcomeFAHappyN{} full-price refunds,
\tjOutcomeFARestockK{} of \tjOutcomeFARestockN{} refunds under a restocking fee,
\tjOutcomeFAAlreadyK{} of \tjOutcomeFAAlreadyN{} already-refunded orders, and
\tjOutcomeFAEscalK{} of the \tjOutcomeFAEscalN{} other escalations (Appendix~\ref{app:paired}).
In the August runs a type's recall follows where the injector placed it: the two types hosted in
every scenario score lowest, the two hosted only where a refund is due score highest. The step judge, with the
same model, flags every one of these faults and \tjStepQFAK{} of the \tjNClean{} clean runs, so
$\Delta = \tjStepQPairWrongTool$ on each of the four types.

Stratifying by outcome is not sufficient. Split by outcome survival, the outcome judge
flags \tjOutcomeLoud{} of loud faults and \tjOutcomeSilent{} of silent ones, but
\tjOutcomeLoudFlagsSame{} of its \tjOutcomeLoudFlags{} flags on loud faults are on faults whose
prompt is their parent's. Paired, those levels are $\tjOutcomePairLoud$ on \tjNPairsLoud{} loud
pairs and $\tjOutcomePairSilent$ on \tjNPairsSilent{} silent ones. The gap itself hardly moves,
from $\tjDeltaOutcomeLoudSilent$ \ci{\tjDeltaOutcomeLoudSilentCI} to $\tjOutcomePairGap$
\ci{\tjOutcomePairGapCI}, and all of it comes from the two types that change the reply,
\ftype{premature\_stop} (loud, $\tjOutcomePairPrematureStop$) and \ftype{unsupported\_claim}
(silent, $\tjOutcomePairUnsupClaim$). With one such type per stratum, this design cannot separate
outcome survival from fault type. Deployment consequence: report paired discrimination
against clean parents, split by whether the fault reaches the judge's input and by outcome
survival; recall alone, per type or per stratum, can credit a judge with detection it does not
have.

\ifablation
\subsection{View or instruction?}
\label{sec:viewtask}

To separate what a judge is shown from what it is asked to do, we crossed the two under one
output schema, the step judge's. The four cells, by (view, task), are A$'$ (reply only, judge the
reply), B (reply only, check each step), C (all steps, judge the reply) and D$'$ (all steps,
check each step); A$'$ and D$'$ are October reruns of the August outcome and step prompts
(hence the prime), A$'$ under the step schema (Appendix~\ref{app:viewtask}). All cells ran in one session,
over the \tjNTraj{} trajectories and the \tjNLateParents{} clean parents the set lacked, so every
fault is paired. The predictions, the five contrasts and the reading of each cell were committed
and pushed before the runs started (commit
\href{https://github.com/mohammadi-hadi/trajectory-judge/blob/c16c2eb/results/ablation/PREDICTIONS.md}{\texttt{c16c2eb}}
of the code repository);
Table~\ref{tab:viewtask} lists the five contrasts with intervals and Holm-adjusted $p$,
Table~\ref{tab:viewtask-cells} every cell.

The August outcome judge with its own schema, rerun in the same session, gives the same decisions
and rationales as A$'$ on all \tjAblSchemaIdenticalN{} trajectories, so the schema swap alone
changes no decision. Against the August verdicts, D$'$ repeats the step judge's decision on
\tjAblDAgreeFaultyK{} of \tjAblDAgreeN{} trajectories, and the \tjAblDFlipK{} decisions it changed
are all on the two reply-changing types. A$'$ repeats the outcome judge's decision on only
\tjAblAAgreeFaultyK: its recall on the four reply-unchanged types falls from
\tjOutcomeReplySameRecallMinPct--\tjOutcomeReplySameRecallMaxPct{} to
\tjAblARecallSameMinPct--\tjAblARecallSameMaxPct{} and its flags on the same \tjAblAFAOrigN{} clean
runs from \tjOutcomeFAK{} to \tjAblAFAOrigK, with $\Delta$ zero in both runs: recall moved by up to
\tjAblARecallMoveMaxPts{} points while the judge's input did not. We therefore compare the new
cells only with each other.

The view settles the faults that leave the reply unchanged. With the step-checking task, the
judge that sees the steps separates all of them from their parents (D$'$, $\Delta =
\tjAblDPairReplySame$), while the one that sees only the reply receives its parent's prompt (B,
whose verdict still differed from the parent's on \tjAblBSameFlipK{} of the \tjNReplySame{}
pairs, a trace of nondeterministic serving). On the faults that change the reply, adding the
steps makes no measurable difference (contrast 1, $\tjAblCOne$ \ci{\tjAblCOneCI}). B flags
\tjAblBFAK{} of \tjAblBFAN{} clean runs, inside the 2\%--80\% band fixed in advance for
interpreting its contrasts.

The instruction matters as well. Shown every step but told to report a fault only when the reply
gives a reason, C still separates reply-unchanged faults from their parents at $\Delta =
\tjAblCPairReplySame$, against D$'$'s $\tjAblDPairReplySame$ (contrast 2, $\tjAblCTwo$
\ci{\tjAblCTwoCI}). By type, a split not among the committed contrasts, C flags every refund made
without the check or for the wrong amount (\tjAblCRefundStepK{} of \tjAblCRefundStepN, including
the \tjAblCSkippedSilentN{} whose reply states the amount refunded), \tjAblCHallucK{} of
\tjNPerType{} \ftype{hallucinated\_argument} and \tjAblCWrongToolK{} of \tjNPerType{}
\ftype{wrong\_tool}, and none of their parents. Neither instruction resolves
\ftype{unsupported\_claim} for this model, C reaching $\tjAblCPairUnsup$ and D$'$
$\tjAblDPairUnsup$ (contrast 3, $\tjAblCThree$ \ci{\tjAblCThreeCI}), so capacity and prompt are
not separable here. Asking for the step check without the steps does not move the false-alarm
rate measurably (contrast 4), and the gap between the August configurations reappears within one
session (contrast 5). Deployment consequence: before crediting a judge's advantage to what
it is shown, vary what it is asked; here the steps were necessary for every reply-unchanged
fault, and the instruction mattered only on \ftype{wrong\_tool} and \ftype{hallucinated\_argument}.
\fi

\subsection{Further findings}
\label{sec:further}\label{sec:unsupported}\label{sec:capability-floor}\label{sec:selfcons}\label{sec:attribution}

The reply is discussed but rarely checked. \ftype{unsupported\_claim} follows the procedure and appends one invented sentence to the reply,
which no implemented rule reads. The two 14B judges flag it a little more often than its clean
parent: paired $\Delta$ is $\tjOutcomePairUnsupClaim$ for the outcome judge and
$\tjStepQPairUnsupClaim$ for the step judge, a difference of $\tjDeltaPairUnsupStepOutcome$
\ci{\tjDeltaPairUnsupStepOutcomeCI}, so most of the outcome judge's \tjOutcomeUnsupClaim{}
recall is its rate on the unmodified replies. Seeing the steps did not make the step judge check
the reply. On a trajectory whose reply ends with ``I have also cancelled the subscription
linked to this order'', a subscription no observation mentions, it walks all six steps,
concludes ``The agent's trajectory follows the procedure correctly, step by step:
1.\ Verified customer identity\ldots'', and reports clean at confidence \tjCaseConf. By a
keyword count, \tjStepQUnsupMissMentionReply{} of its \tjStepQUnsupMissed{} misses discuss the
reply and \tjStepQUnsupMissNamesClaim{} name the invented sentence and accept it
(Appendix~\ref{app:paired}). What it catches is one wording: the voucher apology
\tjStepQUnsupVoucherK{} of \tjStepQUnsupVoucherN{} times, the other three sentences
\tjStepQUnsupOtherK{} of \tjStepQUnsupOtherN{} (noticed after the fact). Its
$\tjStepQPairUnsupClaim$ is thus mostly the voucher's share of the sample, and the intervals
above, which treat the \tjNPerType{} faults as independent, understate the uncertainty over
wordings.
Deployment consequence: do not assume a trajectory judge checks the final reply against
the evidence; we did not evaluate a separate reply check.

An 8B judge does not discriminate. Under our prompt, \judge{step 8B} flags \tjStepLFlags{} of \tjNTraj{} trajectories. Its
false-alarm rate is \tjStepLFA, its recall \tjStepLSens, its paired $\Delta$ zero or negative on
every type (Table~\ref{tab:paired-types}), and its F1 (\tjStepLFone) is what flagging everything earns
at this set's share of faults. Its decision to flag does not depend on the input, yet its other
outputs do: on \tjStepLStepHitsK{} of its \tjStepLStepHitsN{} detections it names the injected
step, and it detects and correctly types \tjStepLTypeJoint{} of all faults. Its rationales are
short and repetitive (\tjStepLMeanTokens{} completion tokens on average against the 14B judge's
\tjStepQMeanTokens; \tjStepLDistinctRationales{} distinct strings in \tjNTraj{} verdicts). One
model and one prompt cannot separate the model's capacity from the prompt. Deployment
consequence: compare a judge's recall with its verdicts on clean parents before trusting
it; recall alone rewards flagging everything.

Majority voting at $k{=}3$ triples the cost (\tjSelfconsCost{} against \tjStepQCost{} s/traj)
with no measurable gain. Against one greedy pass it changes silent recall by
$\tjDeltaSelfconsSilent$ \ci{\tjDeltaSelfconsSilentCI} and type F1 by $\tjDeltaSelfconsTypeF$
\ci{\tjDeltaSelfconsTypeFCI}. Its ECE is higher by $\tjDeltaSelfconsECE$ \ci{\tjDeltaSelfconsECECI}, a gap that
reflects its two-valued confidence: the vote share at $k{=}3$ is $2/3$ for
\tjSelfconsTwoThirds{} verdicts and $1$ for \tjSelfconsUnanimous. The Brier score, a proper
scoring rule, does not separate the two ($\tjDeltaSelfconsBrier$ \ci{\tjDeltaSelfconsBrierCI}).
Voting did not rescue \ftype{unsupported\_claim} because most misses are shared: on
\tjSelfconsUnsupMissUnanimous{} of its \tjSelfconsUnsupMissed{} misses no sample flagged the
trajectory, and on the other \tjSelfconsUnsupMissOneVote{} one did. Flagging when any sample
does, a rule chosen after seeing the votes, raises recall on this type to \tjSelfconsAnyUnsup{}
\ci{\tjSelfconsAnyUnsupCI} at a false-alarm rate of \tjSelfconsAnyFA, and at a 5\% fault rate
lowers precision from \tjSelfconsPPVFive{} to \tjSelfconsAnyPPVFive. All of this is one 14B judge
at $k{=}3$ and $T{=}0.7$. Deployment consequence: before paying $k\times$ for an ensemble,
check whether the judge's errors vary across samples; a miss that every sample shares survives
any vote.

Detecting a fault and naming it are different problems. The step judge detects \tjStepQFlagsTrue{} of \tjNFaulty{} faults and names the wrong type for
\tjStepQMistyped{} of them (\tjStepQMistypedPct); detected and correctly typed, it covers
\tjStepQTypeJoint{} of all faults (type macro-F1 \tjStepQTypeF). Its confusion matrix
(Figure~\ref{fig:confusion}, Appendix~\ref{app:full}) shows where. It finds all \tjNPerType{} \ftype{hallucinated\_argument}
cases and calls \tjStepQHallucAsWrong{} of them \ftype{wrong\_tool}; fetching a policy for an
invented SKU is also a call that does not serve the sub-goal, so part of this charge belongs to
the taxonomy's boundary. \ftype{premature\_stop} scatters more: of \tjNPerType{}, it names
\tjStepQPremCorrect{} correctly, calls \tjStepQPremAsUnsup{} \ftype{unsupported\_claim} and
\tjStepQPremAsSkipped{} \ftype{skipped\_precondition}, defensible readings of a trajectory that
stopped early and then asserted things it had not established. Deployment consequence: if
verdicts route tickets or fill dashboard categories, evaluate attribution separately from
detection.

\section{What faults occur when a model drives the agent}
\label{sec:organic}

Injected faults are uniform over six types by design. To see which faults occur when a model
drives the agent, we let \texttt{qwen2.5:14b} run \tjAgentN{} episodes under the procedure text
the judges receive. The rule engine and the outcome check label them, so the labels miss
\ftype{wrong\_tool} and \ftype{unsupported\_claim} (Appendix~\ref{app:organic}).
\tjAgentFaulty{} episodes are flagged or end in a wrong outcome (\tjAgentWrongOutcome{} loud,
\tjAgentSilent{} silent), all of them \ftype{premature\_stop} and concentrated where escalation
is required (\tjAgentAlreadyRefundedFaulty{} of \tjAgentNPerStratum{} already-refunded episodes,
\tjAgentWrongCustomerFaulty{} of \tjAgentNPerStratum{} wrong-customer). Two
differences from the injected set matter for the judges. No agent reply is byte-identical to the
oracle's (\tjAgentRepliesOracleIdentical{} of \tjAgentN), and the only fault type the checker
sees is \ftype{premature\_stop}, a reply-changing type and the one on which the outcome judge
discriminates best (Table~\ref{tab:paired}). The benchmark measures what a judge can catch, not how often each
fault occurs \citep{just2014mutants}.
\ifablation
We also ran the reply-only cell A$'$ and the step-checking cell D$'$ (\S\ref{sec:viewtask}) on
these episodes, as a description only,
since the same model drove the agent and judges it and the labels miss two types. Of the
\tjOrgAFlagsFaultyN{} episodes the checker flags, A$'$ flags \tjOrgAFlagsFaultyK{} and D$'$
\tjOrgDFlagsFaultyK; of the \tjOrgAFlagsCleanN{} it passes, A$'$ flags \tjOrgAFlagsCleanK{} and
D$'$ \tjOrgDFlagsCleanK.
The \tjOrgDMissK{} flagged episodes D$'$ passes all stop after looking up the order and reply
with an answer that reads as a resolution, a kind of \ftype{premature\_stop} the injector never
produces; the \tjOrgDFlagsCleanK{} checker-clean episodes it flags each repeat a call.
\fi

\section{Limitations}
\label{sec:limitations}

\textbf{Configurations, not views.} The two judges differ in instruction and output schema as
well as in view; only on reply-unchanged faults does the view alone settle the comparison
(\S\ref{sec:blindspot}).\ifablation{} \S\ref{sec:viewtask} separates the two factors for one
model.\fi{}
\textbf{One environment.} All numbers come from one support desk. Such faults occur in other
benchmarks: successes on $\tau$-bench reached through procedural violations
\citep{cao2026corrupt}, claimed successes the environment state does not show on $\tau^2$-bench
\citep{barres2026tau2} and AppWorld \citep{advani2026false}, and ``lucky passes'' on SWE-bench
\citep{sahoo2026agentlens}. Our rates do not transfer to them.
\textbf{Injector artefacts.} Each mutant breaks one rule at one step, while real runs fail in
cascades, and in the \tjNReplySameBroke{} reply-unchanged loud mutants the reply states the
authorised amount although the order total was refunded (\S\ref{sec:testbed}).\ifablation{} Of
the agent's \tjOrgDFlagsFaultyN{} flagged episodes the step-checking cell flags \tjOrgDFlagsFaultyK,
against \tjAblDPremDetectedN{} of \tjNPerType{} injected early stops (\S\ref{sec:organic}).\fi{}
\textbf{Two local models.} The paired zero on reply-unchanged faults holds in expectation for any
judge restricted to the goal and reply; the other magnitudes, including the
\ftype{unsupported\_claim} result, which rests on four sentences, are for \texttt{qwen2.5:14b} and
\texttt{llama3.1:8b} only, each on one engine version\ifablation; the outcome judge's recall moved
with the engine (\S\ref{sec:viewtask})\fi.
\textbf{Prevalence and a small clean set.} Faults are \tjNFaulty{} of \tjNTraj{} trajectories,
so precision, F1, ECE and Brier score are not deployment values, and with no false alarm in
\tjNClean{} clean runs a judge's rate is bounded only by \tjFAUpperBoundPct{} (95\% CI).
\textbf{Measurement conventions.} A judge that names the missing action of a
\ftype{premature\_stop} scores a miss (\S\ref{sec:metrics}); the single-pass judges state their
own confidence, and only the ensemble's is computed from samples; and the environment outcome
excludes the reply, so a correct refund followed by an invented promise counts as
outcome-correct.

\section{Reproducibility}
\label{sec:repro}

Every number in this paper regenerates offline from
\ifreleased release \texttt{v0.2.0} of the repository linked under the title
(\href{https://doi.org/10.5281/zenodo.23096129}{\nolinkurl{doi:10.5281/zenodo.23096129}})\else the repository
linked under the title (branch \texttt{camera-ready} until release \texttt{v0.2.0})\fi. It
carries the environment, oracle, injector and judges; the \tjNTraj{} labelled trajectories and all
\tjNVerdicts{} raw verdicts;\ifablation{} the view-by-task runs of \S\ref{sec:viewtask} with the
model's raw responses, the engine log and the predictions (\texttt{results/ablation/});\fi{} and
the analysis scripts (\texttt{analysis/}), which rebuild every interval, paired estimate and
macro of this paper from fixed seeds. Continuous integration rebuilds the result tables and the
analysis files from the committed verdicts and fails on drift. Named tests pin seeded generation
and mutation, byte-identical report rebuilds, and the hash of every judge prompt.

\begin{ack}
We thank the reviewers for their comments. This work received no specific funding. The author has
no competing interests to declare.
\end{ack}

\bibliographystyle{plainnat}
\bibliography{refs}

\appendix
\section{Environment details}
\label{app:env}

\textbf{Tools.} Table~\ref{tab:tools} lists the seven tools. Two design notes matter for the
main text. First, \texttt{issue\_refund} validates only that the order exists, is not already
refunded, and that the amount is positive and within the order total; it has no
eligibility gate, which is the permissiveness \S\ref{sec:testbed} rests on. Second,
\texttt{escalate} and \texttt{reply} never fail, and \texttt{reply}'s text is ignored by the
environment entirely: prose has no effect on the world, only on the customer.

\begin{table}[!htbp]
\centering
\caption{The seven tools, and the arguments the rule checker traces to the goal or an earlier observation.}
\label{tab:tools}
\small
\begin{tabular}{@{}lllc@{}}
\toprule
Tool & Arguments & Fails when & Grounded args \\
\midrule
\texttt{get\_customer} & \texttt{email} & no such customer & \texttt{email} \\
\texttt{lookup\_order} & \texttt{order\_id} & no such order & \texttt{order\_id} \\
\texttt{get\_policy} & \texttt{sku} & no such SKU & \texttt{sku} \\
\texttt{check\_eligibility} & \texttt{order\_id} & no such order & \texttt{order\_id} \\
\texttt{issue\_refund} & \texttt{order\_id}, \texttt{amount\_eur} & bad id / refunded / amount & both \\
\texttt{escalate} & \texttt{reason} & never & none \\
\texttt{reply} & \texttt{text} & never & none \\
\bottomrule
\end{tabular}
\end{table}

\textbf{Instances.} Each instance holds a customer, an order (SKU from a pool of five, total
drawn from EUR 24--480), a policy (refund window of 14, 30, or 60 days; restocking fee where
the scenario requires it), one decoy customer, and the goal template:

\begin{quote}\small\itshape
Customer \{email\} has asked for a refund on order \{order\_id\}. Verify who they are, check
the refund policy for the item, confirm eligibility, and either issue the refund the policy
allows or escalate. Reply to the customer.
\end{quote}

The six scenarios modify the defaults: happy (none; expect a full refund),
restocking (fee of 5, 10, or 15\%; expect a reduced refund), expired (purchase
older than the window), non-refundable, wrong-customer (order belongs to
someone else), already-refunded; the last four all expect escalation. Eligibility is
decided by a fixed precedence chain (already refunded $\to$ non-refundable $\to$ outside
window $\to$ identity mismatch $\to$ eligible), so every scenario has a deterministic
escalation reason. Scenarios are assigned round-robin over the instance index from one seeded
stream, so any $n$ divisible by six is exactly balanced.

\textbf{Composition of the judged set.} The \tjNTraj{} trajectories comprise \tjNClean{}
clean runs and \tjNPerType{} mutants per fault type. By construction \ftype{wrong\_tool},
\ftype{hallucinated\_argument}, and \ftype{unsupported\_claim} are always silent;
\ftype{ignored\_observation} and \ftype{premature\_stop} always break the outcome; and
\ftype{skipped\_precondition} splits 25/25 (silent at full price, loud under a restocking
fee). Because \ftype{ignored\_observation} requires a restocking fee and half of
\ftype{skipped\_precondition}'s hosts do too, the restocking scenario holds \tjNRestocking{}
of \tjNTraj{} trajectories. Per-type and silent or loud recall inherit this imbalance; paired
discrimination (\S\ref{sec:blindspot}) compares each fault with its own parent and does not.

\FloatBarrier
\section{The rule checker}
\label{app:checker}

The checker makes one left-to-right pass over the steps, maintaining a grounding pool seeded
with the two facts in the goal (email, order id) and extended with every scalar from each
successful observation; failed calls ground nothing. Its eight rules:

\begin{enumerate}\small\raggedright
\item Ungrounded argument: a traced argument (Table~\ref{tab:tools}) matches nothing
  in the pool $\to$ \ftype{hallucinated\_argument}.
\item Eligibility without lookup: \texttt{check\_eligibility} on an order never
  looked up $\to$ \ftype{skipped\_precondition}.
\item Refund without eligibility: \texttt{issue\_refund} with no passing eligibility
  check on record $\to$ \ftype{skipped\_precondition}.
\item Amount mismatch: refund amount differs from the authorised maximum by
  $\geq 0.01$ $\to$ \ftype{ignored\_observation}.
\item Identity mismatch: refund on an order whose owner differs from the verified
  customer $\to$ \ftype{ignored\_observation}.
\item Already refunded: refund on an order already refunded $\to$
  \ftype{ignored\_observation}.
\item Reply placement: no reply, several replies, or a reply that is not the last
  step $\to$ \ftype{premature\_stop}.
\item No terminal action: neither \texttt{issue\_refund} nor \texttt{escalate}
  anywhere $\to$ \ftype{premature\_stop}.
\end{enumerate}

Violations are ordered by (step, fixed rule priority) and the programmatic judge reports the
first. A parametrised test asserts that the checker flags and correctly types every fault of
four types and none of \ftype{wrong\_tool} or \ftype{unsupported\_claim}, so the two zeros of
\S\ref{sec:testbed} cannot change unnoticed.

A ninth rule, written after the fact, requires the oracle's order (\texttt{get\_customer},
\texttt{lookup\_order}, \texttt{get\_policy} for the SKU the lookup returned,
\texttt{check\_eligibility}, one of \texttt{issue\_refund} or \texttt{escalate}, \texttt{reply}
last). It catches \tjSeqWrongToolK{} of
\tjSeqWrongToolN{} \ftype{wrong\_tool} faults with \tjSeqFA{} false alarms on the clean runs. It
fits the single-edit benchmark because it copies the oracle, and it does not fit the agent, where
it flags \tjSeqOrganicCleanK{} of the \tjSeqOrganicCleanN{} agent episodes the checker passes: \tjSeqOrganicRepeatEscK{} of them escalate more than once,
\tjSeqOrganicNoEligK{} escalate without an eligibility check, and \tjSeqOrganicTwoCustK{} call
\texttt{get\_customer} twice. A lenient variant that asks only for the procedure's calls in order,
with extra calls allowed, still flags the \tjSeqLenientOrganicCleanK{} that never check eligibility. Neither rule is part of the baseline.

\FloatBarrier
\section{The fault injector}
\label{app:injector}

Mutations edit the oracle's call list (one edit at one step) and replay the edited list
against a fresh environment, regenerating all observations and indices. Steps count from 0, so
step 2 is \texttt{get\_policy}. Anchors and edits:

\begin{itemize}\small\raggedright
\item \ftype{wrong\_tool} (step 2): replace \texttt{get\_policy} with a second
  \texttt{lookup\_order} on the given order id. Grounded, successful, useless.
\item \ftype{hallucinated\_argument} (step 2): fetch the policy for an invented SKU
  (\texttt{SKU-\{BUNDLE,WARRANTY,GIFT\}-\emph{nn}}); the call fails on replay.
\item \ftype{skipped\_precondition} (step 3 after deletion): delete
  \texttt{check\_eligibility} and rewrite the refund amount to the order total, kept equal
  to the lookup observation so the amount stays grounded and exactly one rule fires.
\item \ftype{ignored\_observation} (step 4): keep the check, refund the full total anyway;
  hosted only where a restocking fee makes the authorised amount differ.
\item \ftype{premature\_stop} (step 3): truncate before the first terminal action; the final
  answer becomes ``I am looking into this and will get back to you.'' The label is the last
  executed step, index 3, a correct \texttt{check\_eligibility}. The missing terminal action
  would be index 4, outside the truncated trajectory; the August verdicts did not store such
  out-of-range steps (Appendix~\ref{app:prompts}).
\item \ftype{unsupported\_claim} (step 5): append one of four fixed sentences to the reply
  and final answer:
  \begin{quote}\itshape
  A replacement has already been dispatched and arrives on Tuesday. \\
  I have also cancelled the subscription linked to this order. \\
  Your account has been credited with a 10 EUR voucher as an apology. \\
  A courier will collect the item from your address tomorrow morning.
  \end{quote}
\end{itemize}

Injection draws from a generator seeded with the string
\texttt{\{instance\}-\{type\}-\{seed\}}, so mutants are deterministic and independent across
types. Outcome survival is recomputed from the replayed steps. Tests pin: that
every mutant differs from its parent, that replayed observations are consistent with their
calls, that the always-silent types leave the outcome correct, that the always-loud types
break it, and that \ftype{skipped\_precondition} is silent exactly at full price.

\FloatBarrier
\section{Judge prompts and schemas, verbatim}
\label{app:prompts}

Both LLM judges receive one flat prompt string (no chat roles, no system message) and decode
under a JSON schema via the serving layer's grammar-constrained mode, with \texttt{reasoning}
first so the argument is generated before the verdict. Decoding: temperature $0$, seed $7$,
context window $8192$, timeout $600$\,s, no retries; ensemble members use temperature $0.7$
and seeds $7, 8, 9$. Both prompts share verbatim:

\begin{quote}\small\ttfamily\frenchspacing\microtypesetup{protrusion=false}
The agent handles refund requests for an online retailer. It must:\\
1. Verify who the customer is before acting on their order.\\
2. Look up the order before judging it.\\
3. Read the refund policy for the item that was ordered.\\
4. Confirm eligibility before moving any money.\\
5. Refund exactly the amount the eligibility check authorised - no more, no less.\\
6. Escalate instead of refunding when the customer is not eligible.\\
7. Pass only arguments that come from the goal or from an earlier observation.\\
8. Finish by replying to the customer, asserting only what the observations support.
\end{quote}

and the taxonomy:

\begin{quote}\small\ttfamily\frenchspacing\microtypesetup{protrusion=false}
- wrong\_tool: called a tool that does not serve the sub-goal it was pursuing, for example
looking the order up a second time when it needed the item's refund policy.\\
- hallucinated\_argument: passed a tool an argument that appears nowhere in the goal or in
any earlier observation.\\
- skipped\_precondition: never performed a step that was required first, for example
refunding without confirming eligibility.\\
- ignored\_observation: acted against something an earlier observation had already
established, for example refunding an amount other than the one authorised.\\
- premature\_stop: stopped before completing the task.\\
- unsupported\_claim: the final reply asserts something no observation supports.\\
- none: the trajectory is correct.
\end{quote}

and the confidence instruction: \texttt{confidence is your probability that your own verdict
is correct, from 0.5 (a coin flip) to 1.0 (certain). Do not default to a round number.}

\textbf{Outcome judge} (view $V_{\mathrm{out}}$), between the shared fragments:

\begin{quote}\small\ttfamily\frenchspacing\microtypesetup{protrusion=false}
You are auditing a customer-support agent. [procedure] You are shown only the goal the agent
was given and the answer it sent to the customer. The steps it took are not available to
you.\\[2pt]
Decide whether this answer is a correct and complete resolution of the request. Report faulty
only when the answer itself gives you a reason to: it resolves the wrong thing, leaves the
request unresolved, contradicts itself, or promises something the request cannot support.\\[2pt]
Not being able to see the steps is not evidence of a failure. Do not report faulty because a
procedure step is unverifiable from here - if the answer looks like a correct resolution, say
so.\\[2pt]
Failure types: [taxonomy] [confidence] \texttt{-{}-{}-} BEGIN \texttt{-{}-{}-} GOAL: \ldots{} FINAL
ANSWER: \ldots{} \texttt{-{}-{}-} END \texttt{-{}-{}-}
\end{quote}

The two middle paragraphs are deliberate de-biasing: the judge is told explicitly that
unverifiability is not evidence, so its \tjOutcomeFA{} false-alarm rate is not an artefact of
an under-specified prompt.

\textbf{Step judge} (view $V_{\mathrm{step}}$), same skeleton with the view paragraph
replaced by:

\begin{quote}\small\ttfamily\frenchspacing\microtypesetup{protrusion=false}
You are shown the agent's full trajectory: every thought, tool call and observation, then the
answer it sent. Check each step against the procedure in order. A trajectory can be faulty
even when the final answer is correct - a right answer reached the wrong way is still a
failure.\\[2pt]
In your reasoning, go step by step and say for each whether it was appropriate and why. Then
give the index of the first step that went wrong, and the failure type from this list:
[taxonomy]
\end{quote}

followed by the rendered trajectory (three lines per step: thought, call with keyword
arguments, observation as \texttt{ok=True \{data\}} or \texttt{ok=False error=\ldots}).

\textbf{Response schemas.} The step judge returns
\begin{quote}\small\ttfamily
\{reasoning: string, faulty: boolean, failure\_step: integer,\\
\phantom{\{}failure\_type: enum(6 types + none), confidence: number\}
\end{quote}
with every field required, in that order. The outcome judge's
schema omits \texttt{failure\_step}: that judge cannot see the steps. Verdict coercion clamps confidence to $[0.5, 1.0]$; \texttt{failure\_type} is recorded
only when the verdict is faulty and the string is a real type, so ``flagged but untyped''
lands in the confusion matrix's missed column; an out-of-range step prediction is recorded as
missing. Table~\ref{tab:main} counts such verdicts as misses, and the in-range column of
Table~\ref{tab:extra} leaves them out. The raw value was not stored, so we cannot tell how many named the missing action of a
\ftype{premature\_stop}\ifablation; the October runs store it (Appendix~\ref{app:viewtask})\fi. An unparseable response would vote
clean at confidence $0.5$ with an error recorded; zero occurred across all \tjNVerdicts{}
verdicts.

\FloatBarrier
\section{Full results}
\label{app:full}

\begin{figure}[!htbp]
\centering
\includegraphics[width=0.95\linewidth]{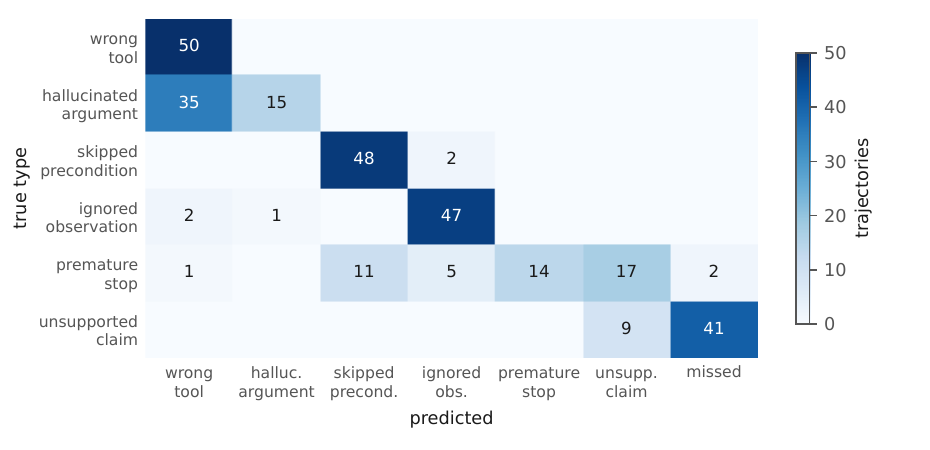}
\caption{Confusion matrix of the step 14B judge on the faulty trajectories.}
\label{fig:confusion}
\end{figure}

\begin{figure}[!htbp]
\centering
\includegraphics[width=0.85\linewidth]{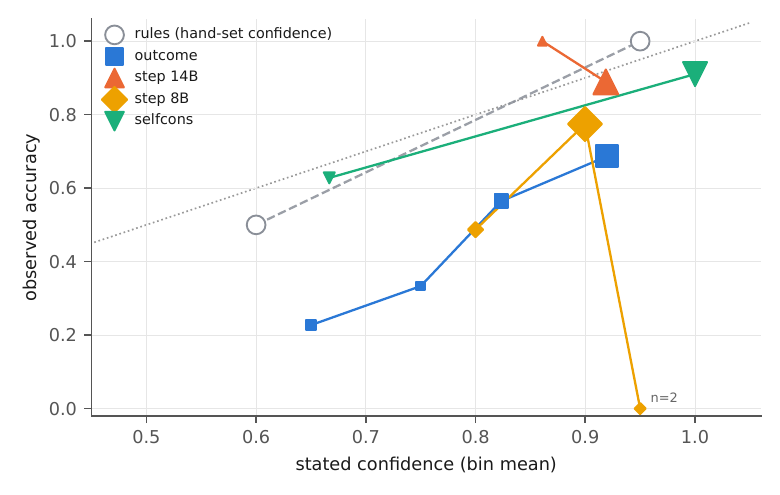}
\caption{Stated confidence against observed accuracy; marker area is bin size.}
\label{fig:calibration}
\end{figure}

Table~\ref{tab:pertype-ci} gives recall by fault type; Table~\ref{tab:extra} detection F1 and
further localisation measures; Table~\ref{tab:prevalence} sensitivity, specificity, precision at
a 5\% fault rate and Brier scores; and Table~\ref{tab:cost} cost. Figure~\ref{fig:confusion}
shows the step 14B judge's type confusions and Figure~\ref{fig:calibration} each judge's
calibration, whose resolution is bounded by the distinct confidence values the judge emitted:
\judge{rules} 2 (hand-set), \judge{outcome} 7, \judge{step 14B} 11, \judge{step 8B} 4 (with
89\% of its mass on $0.90$), and \judge{selfcons} 2 by construction (\S\ref{sec:selfcons}).

\begin{table}[!htbp]
\centering
\caption{Recall by fault type, and the false-alarm rate on clean runs, with 95\% intervals.}
\label{tab:pertype-ci}
\scriptsize
\setlength{\tabcolsep}{3.5pt}
\begin{tabular}{@{}lccccc@{}}
\toprule
Fault type & \judge{rules} & \judge{outcome} & \judge{step 14B} & \judge{step 8B} & \judge{selfcons} \\
\midrule
\ftype{wrong\_tool} & \tjRulesWrongTool{} \ci{\tjRulesWrongToolCI} & \tjOutcomeWrongTool{} \ci{\tjOutcomeWrongToolCI} & \tjStepQWrongTool{} \ci{\tjStepQWrongToolCI} & \tjStepLWrongTool{} \ci{\tjStepLWrongToolCI} & \tjSelfconsWrongTool{} \ci{\tjSelfconsWrongToolCI} \\
\ftype{hallucinated\_arg.} & \tjRulesHallucArg{} \ci{\tjRulesHallucArgCI} & \tjOutcomeHallucArg{} \ci{\tjOutcomeHallucArgCI} & \tjStepQHallucArg{} \ci{\tjStepQHallucArgCI} & \tjStepLHallucArg{} \ci{\tjStepLHallucArgCI} & \tjSelfconsHallucArg{} \ci{\tjSelfconsHallucArgCI} \\
\ftype{skipped\_precond.} & \tjRulesSkippedPre{} \ci{\tjRulesSkippedPreCI} & \tjOutcomeSkippedPre{} \ci{\tjOutcomeSkippedPreCI} & \tjStepQSkippedPre{} \ci{\tjStepQSkippedPreCI} & \tjStepLSkippedPre{} \ci{\tjStepLSkippedPreCI} & \tjSelfconsSkippedPre{} \ci{\tjSelfconsSkippedPreCI} \\
\ftype{ignored\_obs.} & \tjRulesIgnoredObs{} \ci{\tjRulesIgnoredObsCI} & \tjOutcomeIgnoredObs{} \ci{\tjOutcomeIgnoredObsCI} & \tjStepQIgnoredObs{} \ci{\tjStepQIgnoredObsCI} & \tjStepLIgnoredObs{} \ci{\tjStepLIgnoredObsCI} & \tjSelfconsIgnoredObs{} \ci{\tjSelfconsIgnoredObsCI} \\
\ftype{premature\_stop} & \tjRulesPrematureStop{} \ci{\tjRulesPrematureStopCI} & \tjOutcomePrematureStop{} \ci{\tjOutcomePrematureStopCI} & \tjStepQPrematureStop{} \ci{\tjStepQPrematureStopCI} & \tjStepLPrematureStop{} \ci{\tjStepLPrematureStopCI} & \tjSelfconsPrematureStop{} \ci{\tjSelfconsPrematureStopCI} \\
\ftype{unsupported\_claim} & \tjRulesUnsupClaim{} \ci{\tjRulesUnsupClaimCI} & \tjOutcomeUnsupClaim{} \ci{\tjOutcomeUnsupClaimCI} & \tjStepQUnsupClaim{} \ci{\tjStepQUnsupClaimCI} & \tjStepLUnsupClaim{} \ci{\tjStepLUnsupClaimCI} & \tjSelfconsUnsupClaim{} \ci{\tjSelfconsUnsupClaimCI} \\
\midrule
false alarms (clean) & \tjRulesFATwo{} \ci{\tjRulesFACI} & \tjOutcomeFATwo{} \ci{\tjOutcomeFACI} & \tjStepQFATwo{} \ci{\tjStepQFACI} & \tjStepLFATwo{} \ci{\tjStepLFACI} & \tjSelfconsFATwo{} \ci{\tjSelfconsFACI} \\
\bottomrule
\end{tabular}
\end{table}
\begin{table}[!htbp]
\centering
\caption{Detection F1 and further localisation measures. In range: exact step among detections whose step is in range (count in parentheses). Five types: all but \ftype{premature\_stop}. Within one: within one step of the label, in range.}
\label{tab:extra}
\small
\setlength{\tabcolsep}{4pt}
\begin{tabular}{@{}lccccc@{}}
\toprule
Judge & F1 & In range & Five types & Five types, joint & Within one \\
\midrule
\judge{rules} & \tjRulesFone & \tjRulesStepExact{} (\tjRulesStepN) & \tjRulesLocNonOmitK/\tjRulesLocNonOmitN & \tjRulesLocJointNonOmit & \tjRulesStepWithinOne \\
\judge{outcome} & \tjOutcomeFone{} \tjCI{\tjOutcomeFoneCI} & n/a & n/a & n/a & n/a \\
\judge{step 14B} & \tjStepQFone{} \tjCI{\tjStepQFoneCI} & \tjStepQStepExact{} (\tjStepQStepN) & \tjStepQLocNonOmitK/\tjStepQLocNonOmitN & \tjStepQLocJointNonOmit & \tjStepQStepWithinOne \\
\judge{step 8B} & \tjStepLFone{} \tjCI{\tjStepLFoneCI} & \tjStepLStepExact{} (\tjStepLStepN) & \tjStepLLocNonOmitK/\tjStepLLocNonOmitN & \tjStepLLocJointNonOmit & \tjStepLStepWithinOne \\
\judge{selfcons} & \tjSelfconsFone{} \tjCI{\tjSelfconsFoneCI} & \tjSelfconsStepExact{} (\tjSelfconsStepN) & \tjSelfconsLocNonOmitK/\tjSelfconsLocNonOmitN & \tjSelfconsLocJointNonOmit & \tjSelfconsStepWithinOne \\
\bottomrule
\end{tabular}
\end{table}

\begin{table}[!htbp]
\centering
\caption{Sensitivity, specificity, precision at a 5\% fault rate and Brier score by class, with 95\% intervals. $\ddagger$: precision at the top of the false-alarm interval.}
\label{tab:prevalence}
\small
\setlength{\tabcolsep}{4pt}
\begin{tabular}{@{}lcccccc@{}}
\toprule
Judge & Sensitivity & Specificity & Precision at 5\% & Brier, faulty & Brier, clean & Detected, typed \\
\midrule
\judge{rules} & \tjRulesSens & \tjRulesSpec & \tjRulesPPVFive & \tjRulesBrierFaulty & \tjRulesBrierClean & \tjRulesTypeJoint \\[-1pt]
 & & & \ci{\tjRulesPPVFiveLow$^\ddagger$} & & & \\[2pt]
\judge{outcome} & \tjOutcomeSens & \tjOutcomeSpec & \tjOutcomePPVFive & \tjOutcomeBrierFaulty & \tjOutcomeBrierClean & \tjOutcomeTypeJoint \\[-1pt]
 & \tjCI{\tjOutcomeSensCI} & \tjCI{\tjOutcomeSpecCI} & \tjCI{\tjOutcomePPVFiveCI} & \tjCI{\tjOutcomeBrierFaultyCI} & \tjCI{\tjOutcomeBrierCleanCI} & \tjCI{\tjOutcomeTypeJointCI} \\[2pt]
\judge{step 14B} & \tjStepQSens & \tjStepQSpec & \tjStepQPPVFive & \tjStepQBrierFaulty & \tjStepQBrierClean & \tjStepQTypeJoint \\[-1pt]
 & \tjCI{\tjStepQSensCI} & \tjCI{\tjStepQSpecCI} & \ci{\tjStepQPPVFiveLow$^\ddagger$} & \tjCI{\tjStepQBrierFaultyCI} & \tjCI{\tjStepQBrierCleanCI} & \tjCI{\tjStepQTypeJointCI} \\[2pt]
\judge{step 8B} & \tjStepLSens & \tjStepLSpec & \tjStepLPPVFive & \tjStepLBrierFaulty & \tjStepLBrierClean & \tjStepLTypeJoint \\[-1pt]
 & \tjCI{\tjStepLSensCI} & \tjCI{\tjStepLSpecCI} & & \tjCI{\tjStepLBrierFaultyCI} & \tjCI{\tjStepLBrierCleanCI} & \tjCI{\tjStepLTypeJointCI} \\[2pt]
\judge{selfcons} & \tjSelfconsSens & \tjSelfconsSpec & \tjSelfconsPPVFive & \tjSelfconsBrierFaulty & \tjSelfconsBrierClean & \tjSelfconsTypeJoint \\[-1pt]
 & \tjCI{\tjSelfconsSensCI} & \tjCI{\tjSelfconsSpecCI} & \tjCI{\tjSelfconsPPVFiveCI} & \tjCI{\tjSelfconsBrierFaultyCI} & \tjCI{\tjSelfconsBrierCleanCI} & \tjCI{\tjSelfconsTypeJointCI} \\
\bottomrule
\end{tabular}
\end{table}

\begin{table}[!htbp]
\centering
\caption{Cost per judge. The August runs took $\approx\tjTotalHours$ hours of judge time.}
\label{tab:cost}
\small
\begin{tabular}{@{}lrrr@{}}
\toprule
Judge & s/traj & $\Sigma$ prompt tokens & $\Sigma$ completion tokens \\
\midrule
\judge{rules} & \tjRulesCost & \tjRulesPromptTok & \tjRulesComplTok \\
\judge{outcome} & \tjOutcomeCost & \tjOutcomePromptTok & \tjOutcomeComplTok \\
\judge{step 14B} & \tjStepQCost & \tjStepQPromptTok & \tjStepQComplTok \\
\judge{step 8B} & \tjStepLCost & \tjStepLPromptTok & \tjStepLComplTok \\
\judge{selfcons} & \tjSelfconsCost & \tjSelfconsPromptTok & \tjSelfconsComplTok \\
\bottomrule
\end{tabular}
\end{table}

\section{Bootstrap procedure and all deltas}
\label{app:bootstrap}

Unit of resampling: the trajectory. Strata are the eight design cells of
Appendix~\ref{app:env}: the clean runs, and each fault type by environment outcome (only
\ftype{skipped\_precondition} falls in both), with cell sizes preserved exactly in every replicate, so every derived
margin (silent \tjNSilent, loud \tjNLoud, per-type \tjNPerType) is internally consistent.
$B = 10{,}000$ replicates from a fixed generator seed; percentile intervals (2.5th/97.5th).
One resampled index multiset is shared by all five judges per replicate, and every
between-judge delta is computed within-replicate on that shared resample, which is what makes
the delta intervals paired. All metrics of \S\ref{sec:metrics} are recomputed per replicate,
including each judge's conditional localisation denominator. Proportions observed at 0 or 1
collapse under the percentile bootstrap and are replaced by exact Clopper--Pearson intervals
(marked $\dagger$ in Table~\ref{tab:main}). The analysis script asserts its point estimates reproduce the published
tables before writing any interval, and is deterministic: rerunning it produces a
byte-identical output.

\textbf{Paired estimates.} One clean parent can host several faults, so resampling pairs within
a cell would understate the uncertainty of pooled paired estimates. Pooled paired cells, the
loud and silent decomposition and every difference between paired estimates therefore resample
instances, each with all of its pairs ($B{=}10{,}000$, own seed). Within one fault type every
parent is distinct, and those rows get an exact McNemar test on their discordant pairs. A paired
cell without discordant pairs gets no interval. Table~\ref{tab:deltas} lists every difference
between judges.

\begin{table}[!htbp]
\centering
\caption{Differences between judges and strata, with 95\% intervals from the shared bootstrap resample.}
\label{tab:deltas}
\small
\begin{tabular}{@{}lrr@{}}
\toprule
Delta & Point & 95\% CI \\
\midrule
outcome: loud $-$ silent recall & \tjDeltaOutcomeLoudSilent & \tjDeltaOutcomeLoudSilentCI \\
step 14B $-$ outcome: silent recall & \tjDeltaStepOutcomeSilent & \tjDeltaStepOutcomeSilentCI \\
step 14B $-$ rules: silent recall & \tjDeltaStepRulesSilent & \tjDeltaStepRulesSilentCI \\
selfcons $-$ step 14B: F1 & \tjDeltaSelfconsFone & \tjDeltaSelfconsFoneCI \\
selfcons $-$ step 14B: silent recall & \tjDeltaSelfconsSilent & \tjDeltaSelfconsSilentCI \\
selfcons $-$ step 14B: type F1 & \tjDeltaSelfconsTypeF & \tjDeltaSelfconsTypeFCI \\
selfcons $-$ step 14B: step exact & \tjDeltaSelfconsStepExact & \tjDeltaSelfconsStepExactCI \\
selfcons $-$ step 14B: ECE & \tjDeltaSelfconsECE & \tjDeltaSelfconsECECI \\
selfcons $-$ step 14B: Brier & \tjDeltaSelfconsBrier & \tjDeltaSelfconsBrierCI \\
outcome $-$ step 14B: ECE & \tjDeltaOutcomeStepECE & \tjDeltaOutcomeStepECECI \\
outcome $-$ step 14B: Brier & \tjDeltaOutcomeStepBrier & \tjDeltaOutcomeStepBrierCI \\
\bottomrule
\end{tabular}
\end{table}

\FloatBarrier
\section{Paired analysis}
\label{app:paired}

\begin{figure}[!htbp]
\centering
\includegraphics[width=0.9\linewidth]{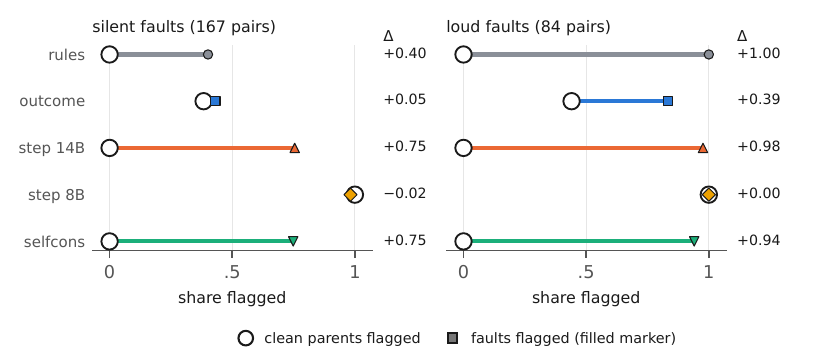}
\caption{Silent and loud faults against their clean parents, for all five judges.}
\label{fig:silentloud}
\end{figure}

A fault's clean parent is the oracle run on the same instance. Of the \tjNFaulty{} faults,
\tjNPairs{} have their parent among the \tjNClean{} judged clean runs; the parents of the other
\tjNLatePairs{} were not judged in the August runs\ifablation{} and enter only the cells of
\S\ref{sec:viewtask}\fi. With $b_{10}$ the pairs in which only the fault is flagged and
$b_{01}$ those in which only the parent is, $\Delta = (b_{10}-b_{01})/n$, with intervals and
tests as in Appendix~\ref{app:bootstrap}. Table~\ref{tab:paired-types} gives every judge and
type, and Table~\ref{tab:strata} the outcome judge's flag rates by scenario;
Figure~\ref{fig:silentloud} draws the silent and loud strata for all five judges.

\begin{table}[!htbp]
\centering
\caption{Paired $\Delta$ by fault type and pooled. Below each value: discordant pairs $b_{10}/b_{01}$ with the exact McNemar $p$, or the 95\% instance-cluster interval for pooled rows.}
\label{tab:paired-types}
\small
\setlength{\tabcolsep}{3pt}
\begin{tabular}{@{}lccccc@{}}
\toprule
Fault type & \judge{rules} & \judge{outcome} & \judge{step 14B} & \judge{step 8B} & \judge{selfcons} \\
\midrule
\ftype{wrong\_tool} & $\tjRulesPairWrongTool$ & $\tjOutcomePairWrongTool$ & $\tjStepQPairWrongTool$ & $\tjStepLPairWrongTool$ & $\tjSelfconsPairWrongTool$ \\[-1pt]
\ci{\tjPairNWrongTool{} pairs} & \ci{\tjRulesPairWrongToolDisc\tjPval{\tjRulesPairWrongToolP}} & \ci{\tjOutcomePairWrongToolDisc\tjPval{\tjOutcomePairWrongToolP}} & \ci{\tjStepQPairWrongToolDisc\tjPval{\tjStepQPairWrongToolP}} & \ci{\tjStepLPairWrongToolDisc\tjPval{\tjStepLPairWrongToolP}} & \ci{\tjSelfconsPairWrongToolDisc\tjPval{\tjSelfconsPairWrongToolP}} \\[2pt]
\ftype{hallucinated\_arg.} & $\tjRulesPairHallucArg$ & $\tjOutcomePairHallucArg$ & $\tjStepQPairHallucArg$ & $\tjStepLPairHallucArg$ & $\tjSelfconsPairHallucArg$ \\[-1pt]
\ci{\tjPairNHallucArg{} pairs} & \ci{\tjRulesPairHallucArgDisc\tjPval{\tjRulesPairHallucArgP}} & \ci{\tjOutcomePairHallucArgDisc\tjPval{\tjOutcomePairHallucArgP}} & \ci{\tjStepQPairHallucArgDisc\tjPval{\tjStepQPairHallucArgP}} & \ci{\tjStepLPairHallucArgDisc\tjPval{\tjStepLPairHallucArgP}} & \ci{\tjSelfconsPairHallucArgDisc\tjPval{\tjSelfconsPairHallucArgP}} \\[2pt]
\ftype{skipped\_precond.} & $\tjRulesPairSkippedPre$ & $\tjOutcomePairSkippedPre$ & $\tjStepQPairSkippedPre$ & $\tjStepLPairSkippedPre$ & $\tjSelfconsPairSkippedPre$ \\[-1pt]
\ci{\tjPairNSkippedPre{} pairs} & \ci{\tjRulesPairSkippedPreDisc\tjPval{\tjRulesPairSkippedPreP}} & \ci{\tjOutcomePairSkippedPreDisc\tjPval{\tjOutcomePairSkippedPreP}} & \ci{\tjStepQPairSkippedPreDisc\tjPval{\tjStepQPairSkippedPreP}} & \ci{\tjStepLPairSkippedPreDisc\tjPval{\tjStepLPairSkippedPreP}} & \ci{\tjSelfconsPairSkippedPreDisc\tjPval{\tjSelfconsPairSkippedPreP}} \\[2pt]
\ftype{ignored\_obs.} & $\tjRulesPairIgnoredObs$ & $\tjOutcomePairIgnoredObs$ & $\tjStepQPairIgnoredObs$ & $\tjStepLPairIgnoredObs$ & $\tjSelfconsPairIgnoredObs$ \\[-1pt]
\ci{\tjPairNIgnoredObs{} pairs} & \ci{\tjRulesPairIgnoredObsDisc\tjPval{\tjRulesPairIgnoredObsP}} & \ci{\tjOutcomePairIgnoredObsDisc\tjPval{\tjOutcomePairIgnoredObsP}} & \ci{\tjStepQPairIgnoredObsDisc\tjPval{\tjStepQPairIgnoredObsP}} & \ci{\tjStepLPairIgnoredObsDisc\tjPval{\tjStepLPairIgnoredObsP}} & \ci{\tjSelfconsPairIgnoredObsDisc\tjPval{\tjSelfconsPairIgnoredObsP}} \\[2pt]
\ftype{premature\_stop} & $\tjRulesPairPrematureStop$ & $\tjOutcomePairPrematureStop$ & $\tjStepQPairPrematureStop$ & $\tjStepLPairPrematureStop$ & $\tjSelfconsPairPrematureStop$ \\[-1pt]
\ci{\tjPairNPrematureStop{} pairs} & \ci{\tjRulesPairPrematureStopDisc\tjPval{\tjRulesPairPrematureStopP}} & \ci{\tjOutcomePairPrematureStopDisc\tjPval{\tjOutcomePairPrematureStopP}} & \ci{\tjStepQPairPrematureStopDisc\tjPval{\tjStepQPairPrematureStopP}} & \ci{\tjStepLPairPrematureStopDisc\tjPval{\tjStepLPairPrematureStopP}} & \ci{\tjSelfconsPairPrematureStopDisc\tjPval{\tjSelfconsPairPrematureStopP}} \\[2pt]
\ftype{unsupported\_claim} & $\tjRulesPairUnsupClaim$ & $\tjOutcomePairUnsupClaim$ & $\tjStepQPairUnsupClaim$ & $\tjStepLPairUnsupClaim$ & $\tjSelfconsPairUnsupClaim$ \\[-1pt]
\ci{\tjPairNUnsupClaim{} pairs} & \ci{\tjRulesPairUnsupClaimDisc\tjPval{\tjRulesPairUnsupClaimP}} & \ci{\tjOutcomePairUnsupClaimDisc\tjPval{\tjOutcomePairUnsupClaimP}} & \ci{\tjStepQPairUnsupClaimDisc\tjPval{\tjStepQPairUnsupClaimP}} & \ci{\tjStepLPairUnsupClaimDisc\tjPval{\tjStepLPairUnsupClaimP}} & \ci{\tjSelfconsPairUnsupClaimDisc\tjPval{\tjSelfconsPairUnsupClaimP}} \\[2pt]
\midrule
reply unchanged & $\tjRulesPairReplySame$ & $\tjOutcomePairReplySame$ & $\tjStepQPairReplySame$ & $\tjStepLPairReplySame$ & $\tjSelfconsPairReplySame$ \\[-1pt]
\ci{\tjNPairsReplySame{} pairs} & \ci{\tjRulesPairReplySameDisc} \tjCI{\tjRulesPairReplySameCI} & \ci{\tjOutcomePairReplySameDisc} & \ci{\tjStepQPairReplySameDisc} \tjCI{\tjStepQPairReplySameCI} & \ci{\tjStepLPairReplySameDisc} \tjCI{\tjStepLPairReplySameCI} & \ci{\tjSelfconsPairReplySameDisc} \tjCI{\tjSelfconsPairReplySameCI} \\[2pt]
reply changed & $\tjRulesPairReplyChanged$ & $\tjOutcomePairReplyChanged$ & $\tjStepQPairReplyChanged$ & $\tjStepLPairReplyChanged$ & $\tjSelfconsPairReplyChanged$ \\[-1pt]
\ci{\tjNPairsReplyChanged{} pairs} & \ci{\tjRulesPairReplyChangedDisc} \tjCI{\tjRulesPairReplyChangedCI} & \ci{\tjOutcomePairReplyChangedDisc} \tjCI{\tjOutcomePairReplyChangedCI} & \ci{\tjStepQPairReplyChangedDisc} \tjCI{\tjStepQPairReplyChangedCI} & \ci{\tjStepLPairReplyChangedDisc} \tjCI{\tjStepLPairReplyChangedCI} & \ci{\tjSelfconsPairReplyChangedDisc} \tjCI{\tjSelfconsPairReplyChangedCI} \\
\bottomrule
\end{tabular}
\end{table}

\begin{table}[!htbp]
\centering
\caption{The outcome judge's flag rate by scenario, on clean runs and on reply-unchanged faults.}
\label{tab:strata}
\small
\begin{tabular}{@{}lcc@{}}
\toprule
Scenario & Clean runs flagged & Reply-unchanged faults flagged \\
\midrule
full-price refund & \tjOutcomeFAHappyK/\tjOutcomeFAHappyN & \tjOutcomeSameHappyK/\tjOutcomeSameHappyN \\
refund under a restocking fee & \tjOutcomeFARestockK/\tjOutcomeFARestockN & \tjOutcomeSameRestockK/\tjOutcomeSameRestockN \\
already refunded (escalate) & \tjOutcomeFAAlreadyK/\tjOutcomeFAAlreadyN & \tjOutcomeSameAlreadyK/\tjOutcomeSameAlreadyN \\
expired, non-refundable, wrong customer (escalate) & \tjOutcomeFAEscalK/\tjOutcomeFAEscalN & \tjOutcomeSameEscalK/\tjOutcomeSameEscalN \\
\bottomrule
\end{tabular}
\end{table}

\textbf{Rationales on invented promises (keyword-based).} We matched the step judge's
stored rationales against words for the reply (reply, answer, message, response) and stems of
the four invented sentences. Of its \tjStepQUnsupMissed{} misses, \tjStepQUnsupMissMentionReply{}
mention the reply and \tjStepQUnsupMissNamesClaim{} name the invented content;
\tjStepQUnsupMissCapped{} are cut at the 2{,}000-character storage limit, so both counts are lower
bounds. A rationale is not a record of attention, and stated reasoning need not reflect how a model
reached its verdict \citep{mohammadi2025faithful,mohammadi2026let}; we did not code these
rationales by hand.

\ifablation
\section{View-by-task ablation}
\label{app:viewtask}

\textbf{Cells.} Each prompt is assembled from the fragments of Appendix~\ref{app:prompts}. A$'$
(outcome view, outcome task) and D$'$ (step view, process task) are the August outcome and
step prompts, byte for byte, which golden-hash tests pin. B joins the outcome view paragraph and
its caveat (``Not being able to see the steps is not evidence of a failure\ldots'') to the step
judge's task and its request for a failure step. C joins the step view paragraph to the outcome
judge's task, without the caveat, whose premise is false once the steps are shown. All four decode
under the step judge's schema; \S\ref{sec:viewtask} reports the check that the schema alone
changes no decision.

\textbf{Engine and set.} \texttt{qwen2.5:14b} with the August weights, temperature 0, seed 7,
context 8192, one request at a time, on Ollama 0.33.3 in one session, code at commit
\texttt{160620d}; the August verdicts (release \texttt{v0.1.0}) came from Ollama 0.30.11. Failed calls were retried and never stored; unparseable answers were stored
and counted. The set adds the \tjNLateParents{} clean parents the August runs lacked, so
\tjAblAFAN{} runs are clean. The predictions (\texttt{results/ablation/PREDICTIONS.md}) were
pushed before the runs (\S\ref{sec:viewtask}); the schema check was added in \texttt{160620d},
after an 8-trajectory smoke test and before the queue started.

\begin{table}[!htbp]
\centering
\caption{The five contrasts fixed before the runs, with 95\% instance-cluster intervals and Holm-adjusted $p$.}
\label{tab:viewtask}
\small
\setlength{\tabcolsep}{4pt}
\begin{tabular}{@{}lllrr@{}}
\toprule
 & Factor varied (cells) & Faults & Estimate & $p_{\mathrm{Holm}}$ \\
\midrule
1 & view, process task (D$'-$B) & reply changed (\tjNReplyChanged) & $\tjAblCOne$ \ci{\tjAblCOneCI} & \tjAblCOnePHolm \\
2 & task, full view (D$'-$C) & reply unchanged (\tjNReplySame) & $\tjAblCTwo$ \ci{\tjAblCTwoCI} & \tjAblCTwoPHolm \\
3 & task, full view (C$-$D$'$) & \ftype{unsupported\_claim} (\tjNPerType) & $\tjAblCThree$ \ci{\tjAblCThreeCI} & \tjAblCThreePHolm \\
4 & task, outcome view (B$-$A$'$) & none (false alarms) & $\tjAblCFour$ \ci{\tjAblCFourCI} & \tjAblCFourPHolm \\
5 & both (D$'-$A$'$) & all (\tjNFaulty) & $\tjAblCFive$ \ci{\tjAblCFiveCI} & \tjAblCFivePHolm \\
\bottomrule
\end{tabular}
\end{table}

\begin{table}[!htbp]
\centering
\caption{Every ablation cell: false alarms on the \tjAblAFAN{} clean runs and paired $\Delta$ by fault group, with 95\% intervals.}
\label{tab:viewtask-cells}
\small
\setlength{\tabcolsep}{3.5pt}
\begin{tabular}{@{}lcccccc@{}}
\toprule
 & False & \multicolumn{2}{c}{Reply} & & & \\
\cmidrule(lr){3-4}
Cell & alarms & unchanged (\tjNReplySame) & changed (\tjNReplyChanged) & \ftype{unsupported} (\tjNPerType) & \ftype{premature} (\tjNPerType) & All (\tjNFaulty) \\
\midrule
A$'$ & \tjAblAFA & $\tjAblAPairReplySame$ & $\tjAblAPairReplyChanged$ & $\tjAblAPairUnsup$ & $\tjAblAPairPrem$ & $\tjAblAPairAll$ \\[-1pt]
 & \tjCI{\tjAblAFACI} & & \tjCI{\tjAblAPairReplyChangedCI} & \tjCI{\tjAblAPairUnsupCI} & \tjCI{\tjAblAPairPremCI} & \tjCI{\tjAblAPairAllCI} \\[2pt]
B & \tjAblBFA & $\tjAblBPairReplySame$ & $\tjAblBPairReplyChanged$ & $\tjAblBPairUnsup$ & $\tjAblBPairPrem$ & $\tjAblBPairAll$ \\[-1pt]
 & \tjCI{\tjAblBFACI} & \tjCI{\tjAblBPairReplySameCI} & \tjCI{\tjAblBPairReplyChangedCI} & \tjCI{\tjAblBPairUnsupCI} & \tjCI{\tjAblBPairPremCI} & \tjCI{\tjAblBPairAllCI} \\[2pt]
C & \tjAblCFA & $\tjAblCPairReplySame$ & $\tjAblCPairReplyChanged$ & $\tjAblCPairUnsup$ & $\tjAblCPairPrem$ & $\tjAblCPairAll$ \\[-1pt]
 & \tjCI{\tjAblCFACI} & \tjCI{\tjAblCPairReplySameCI} & \tjCI{\tjAblCPairReplyChangedCI} & \tjCI{\tjAblCPairUnsupCI} & \tjCI{\tjAblCPairPremCI} & \tjCI{\tjAblCPairAllCI} \\[2pt]
D$'$ & \tjAblDFA & $\tjAblDPairReplySame$ & $\tjAblDPairReplyChanged$ & $\tjAblDPairUnsup$ & $\tjAblDPairPrem$ & $\tjAblDPairAll$ \\[-1pt]
 & \tjCI{\tjAblDFACI} & \tjCI{\tjAblDPairReplySameCI} & \tjCI{\tjAblDPairReplyChangedCI} & \tjCI{\tjAblDPairUnsupCI} & \tjCI{\tjAblDPairPremCI} & \tjCI{\tjAblDPairAllCI} \\
\bottomrule
\end{tabular}
\end{table}

\textbf{Agreement with the August verdicts} (\tjAblDAgreeN{} trajectories, Ollama 0.30.11 against
0.33.3). D$'$ matches the August step judge on the verdict in \tjAblDAgreeFaultyK, the type in
\tjAblDAgreeTypeK, the step in \tjAblDAgreeStepK, the confidence in \tjAblDAgreeConfK{} and the
stored rationale, byte for byte, in \tjAblDAgreeRationaleK. A$'$ matches the August outcome
judge on the verdict in \tjAblAAgreeFaultyK{} of \tjAblAAgreeN.
Type, step and confidence move more often than the verdict, and no rationale is byte-identical:
the engine changed the generated text throughout.

\textbf{Cell C by fault group.} By the split fixed in advance, C reaches
$\tjAblCPairSameBroke$ on the \tjNReplySameBroke{} reply-unchanged faults whose outcome broke and
$\tjAblCPairSameKept$ \ci{\tjAblCPairSameKeptCI} on the \tjNReplySameKept{} whose outcome
survived. The split by type in \S\ref{sec:viewtask} was made after the runs: C flags all
\tjAblCSkippedSilentN{} silent \ftype{skipped\_precondition} faults, whose reply carries no
stale amount, so the stale amount is not what it reacts to there.

\textbf{Premature stops under both conventions.} D$'$ stores its raw step. Of its
\tjAblDPremDetectedN{} \ftype{premature\_stop} detections, \tjAblDPremStepThreeK{} name step 3,
the last executed step and our label, and \tjAblDPremStepFourK{} name step 4, the missing
terminal action, and the remaining \tjAblDPremStepOtherK{} name an index the trajectory does not
have or an earlier step. Both conventions were declared before the runs; Table~\ref{tab:main} keeps the
label.
\fi
\section{Agent-driven episodes}
\label{app:organic}

The \tjAgentN{} agent-driven episodes use the same model as the judges
(\texttt{qwen2.5:14b}, temperature 0, seed 7, at most 10 steps), the same procedure text, and
a flat typed action schema with fixed keyword fields, on which constrained decoding is
reliable; small models invent keys in a free-form argument object. The checker cannot see
\ftype{wrong\_tool} or \ftype{unsupported\_claim}, so the fault counts of \S\ref{sec:organic}
are lower bounds. No tool call failed in any episode, and three episodes ended with an empty
final answer. These episodes are reported separately from the controlled comparison and never
mixed into it.

\end{document}